\documentclass[letterpaper]{article} % DO NOT CHANGE THIS
\usepackage[preprint]{aaai2027}

\usepackage[hyphens]{url}  % DO NOT CHANGE THIS
\usepackage{graphicx} % DO NOT CHANGE THIS
\usepackage{natbib}  % DO NOT CHANGE THIS AND DO NOT ADD ANY OPTIONS TO IT
\usepackage{caption} % DO NOT CHANGE THIS AND DO NOT ADD ANY OPTIONS TO IT
\usepackage{amsmath}  % equations: cases, aligned, \text
\usepackage{amssymb}  % \mathbb
\usepackage{booktabs} % compact, publication-quality tables
\usepackage{array}    % ragged-right fixed-width table columns
\usepackage{cuted}
\usepackage{dblfloatfix}
\usepackage{algorithm}
\usepackage{algorithmic}
\newcolumntype{L}[1]{>{\raggedright\arraybackslash}p{#1}}
\title{CheckVLA: Execution-Time Verification with Action-Conditioned World Model
\\for Long-Horizon Mobile Manipulation}
\author{
    Yushan Liu\textsuperscript{\rm 1},
    Peibo Sun\textsuperscript{\rm 2},
    Xintao Chao\textsuperscript{\rm 1},
    Zhenyang Yang\textsuperscript{\rm 3},
    Yifan Xie\textsuperscript{\rm 1},
    Lingfeng Zhang\textsuperscript{\rm 1}, \\
    Shoujie Li\textsuperscript{\rm 4},
    Chenyu Tang\textsuperscript{\rm 3},
    Fang Chen\textsuperscript{\rm 2},
    Xiao-Ping Zhang\textsuperscript{\rm 1},
    Wenbo Ding\textsuperscript{\rm 1,5}\corresponding
}
\affiliations{
    \textsuperscript{\rm 1}Tsinghua University \quad
    \textsuperscript{\rm 2}Shanghai Jiao Tong University \quad
    \textsuperscript{\rm 3}Peking University \\
    \textsuperscript{\rm 4}Nanyang Technological University \quad
    \textsuperscript{\rm 5}Xspark AI
}

\begin{document}

\maketitle

\begin{abstract}

Vision-language-action (VLA) policies commonly execute long-horizon mobile manipulation through open-loop action chunks, issuing multiple actions without receiving new high-level visual input. A committed chunk therefore implies how observations should evolve, but accidental deviations can violate this expectation while the remaining actions continue to propagate the error: commit-time policy confidence cannot react to a deviation that occurs after dispatch, and observation-only anomaly scores lack an action-conditioned reference for separating expected effects from unexplained changes. We propose \textbf{CheckVLA}, which verifies execution with a separately trained, frozen action-conditioned world model. A conformally calibrated risk threshold bounds the episode-level probability of an unnecessary first intervention and determines when to intervene, its exceedance controls how strongly the rewritten suffix retains the superseded chunk, latency-aware hard prefixing restricts replacement to actions that remain deployable, and an event-driven keyframe bank preserves evidence of prior progress across repairs. On RoboCasa365, under a common training recipe and a matched invocation budget, CheckVLA attains a 36.1\% average success rate against 27.6\% for periodic replanning (+8.5 points). At a matched 5\% episode-level false-alarm target, action conditioning raises timely recall to 77.9\%, against 48.6\% for an observation-only control and 37.9\% for an action-shuffled control. These simulation results support action-conditioned verification as a way to restore feedback during chunked execution while keeping the repair consistent with inference latency.

\end{abstract}

\section{Introduction}

Long-horizon mobile manipulation interleaves navigation with contact-rich interaction, making it a demanding testbed for embodied intelligence. Recent VLA systems support mobile whole-body control across increasingly diverse tasks \cite{pi05,mem,pi07,abot_m05,wholebodyvla,future_vla}. Many execute a fixed-length action chunk without re-querying the policy on updated observations \cite{act,diffusion_policy,pi0}. Chunking amortizes inference and improves temporal consistency, but each issued chunk also predicts that observations will evolve consistently with its committed actions.

A post-dispatch deviation can violate this prediction without prompting replanning: an object may slip while subsequent actions still assume a secure grasp, or a displaced base may continue smooth but incorrect motion. Because the remaining actions can stay locally plausible after their premise fails, commit-time policy scores may miss this \emph{confidently wrong} regime. The resulting errors compound over long horizons \cite{ross2010,dagger,three_regimes}, motivating an execution-time test of whether committed actions produce their expected consequences.

Existing signals cover only parts of this problem. Commit-time uncertainty \cite{bid,moh,autohorizon} cannot use observations arriving after dispatch without another policy call. Post-dispatch monitors use observations, policy features, or world-model scores \cite{reflect,doremi,safe,foresight,gauge}, but observation-only variants lack an action-conditioned reference. Adaptive execution and asynchronous continuity methods \cite{pistar06,pi07,trust_imagination,rtc,remac} address when to refresh a chunk or how to preserve continuity, yet do not generally combine an episode-calibrated online test with latency-aware suffix repair. This leaves open how to turn action-conditioned consequence predictions into a calibrated intervention and a deployable correction.

\begin{figure*}[t]
\centering
\includegraphics[width=\textwidth]{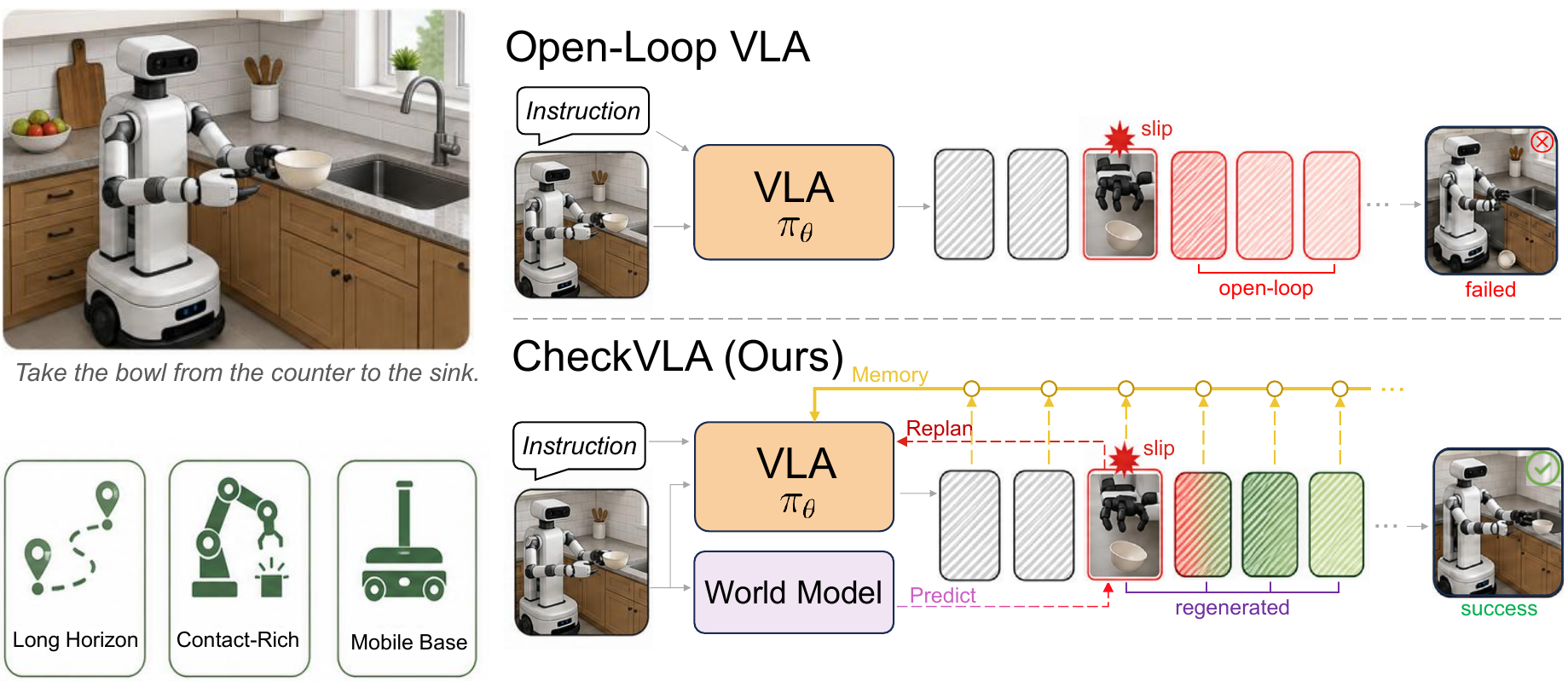}
\caption{Overview of CheckVLA. Open-loop chunks can propagate in-chunk failures such as slippage. CheckVLA compares action-conditioned predictions with arriving observations, uses an episode-calibrated threshold to trigger intervention, and rewrites the repairable suffix while preserving episodic context.}
\label{fig:overview}
\vskip -0.2in
\end{figure*}

We present \textbf{CheckVLA}, which tests committed actions against observed consequences during execution (Figure~\ref{fig:overview}). A separately trained, frozen world model predicts short-horizon features from the latest observation and remaining actions; a causal risk head aggregates prediction--observation discrepancies. On exchangeable nominal successes, a functional conformal threshold bounds the probability of an unnecessary first intervention. The first crossing sets repair timing, while a validation-selected exceedance map sets reference retention. The same VLA rewrites only the latency-feasible suffix, and a keyframe bank preserves episodic context.

% Keep this figure* declaration anchored after a full paragraph (not directly
% after a heading): declared here it lands on page 2 top; declared immediately
% after \section{Introduction} the dblfloatfix placement degrades (fig -> p3).

On RoboCasa365 \cite{robocasa365}, CheckVLA achieves \textbf{36.1\%} average success, setting the state of the art among published Human300 policy-pretraining results (Table~\ref{tab:robocasa365_flagships}). Under a common internal training recipe and paired seeds, it exceeds invocation-matched periodic replanning by 8.5 percentage points. At a common target episode-level family-wise error rate (FWER), removing action conditioning reduces timely recall from 77.9\% to 48.6\% for an observation-only control and 37.9\% for an action-shuffled control.

Our contributions are as follows:
\begin{itemize}
\item We cast an open-loop action chunk as a testable prediction of near-future observations, linking failure detection to action-conditioned consequence prediction and repair to rewriting the latency-feasible suffix.
\item We develop CheckVLA with an episode-calibrated trigger, validation-selected risk-adaptive retention, a latency-aware hard prefix, and event-driven episodic memory.
\item CheckVLA reaches state-of-the-art average success among RoboCasa365 Human300 policy-pretraining methods and exceeds invocation-matched periodic replanning under a common recipe; controlled studies audit detection, repair, memory, latency, robustness, runtime, and failures.
\end{itemize}

\section{Related Work}

\subsection{VLA for Mobile Manipulation}

VLA models jointly predict base, arm, and gripper actions for mobile platforms \cite{mobile_aloha,pi0,pi05,wholebodyvla,avr}; cross-embodiment training, subtask decoding, video memory, and action-subspace decoupling broaden their coverage \cite{pi05,mem,pi07,abot_m05,legacy_transfer,exoviha}. Because these systems commonly commit fixed-length chunks \cite{act,diffusion_policy}, long-horizon extensions have mainly changed training or boundary-time context \cite{pistar06,abot_m05,pi07}. CheckVLA instead verifies a chunk during execution and rewrites its remaining suffix.

\subsection{World Models for Robotics}

Action-conditioned world models support policy conditioning, data generation, and planning \cite{dreamvla,worldvla,dino_wm,vjepa2,perceptdrive}; several predict frozen-space features without pixel decoding \cite{dino_wm,oawam}, building on latent imagination \cite{planet,dreamer,skyjepa}. Related systems condition a VLA on predicted features \cite{vla_jepa}, couple imagination to action generation \cite{abot_m05,forgedrive}, verify world-model predictions \cite{world_action_verifier}, or adapt execution length by comparing predicted and realized futures \cite{trust_imagination}. CheckVLA uses consequence predictions as a calibrated execution-time test, separates the verifier from the policy path, and converts an alarm into latency-aware suffix repair.

\subsection{Execution Monitoring and Intervention}

Adaptive chunking changes the execution window using latency or policy-side uncertainty and preserves continuity under asynchronous inference \cite{rtc,bid,moh,autohorizon,remac}. Monitoring uses constraints, generated checks, policy features, or world-model scores \cite{doremi,code_as_monitor,safe,foresight,gauge}, while pre-dispatch screening, backup policies, guided regeneration, and keyframe memory address complementary stages \cite{pre_vla,aegis,vla_corrector,kemo,mem}. CheckVLA combines an episode-calibrated action-conditioned trigger with a validation-selected exceedance map for latency-feasible repair and persistent episodic context; Table~\ref{tab:capability} positions it against the closest execution-time systems.

\begin{table}[t]
\centering
\small
\setlength{\tabcolsep}{1.2pt}
\begin{tabular*}{\columnwidth}{@{\extracolsep{\fill}}lccccc@{}}
\toprule
Method
& \shortstack{Act.-cond.\\signal}
& \shortstack{Calib.\\trigger}
& \shortstack{In-chunk\\repair}
& \shortstack{Latency\\aware}
& \shortstack{Epis.\\memory} \\
\midrule
SAFE & -- & \checkmark & -- & -- & -- \\
Foresight & \checkmark & \checkmark & -- & -- & -- \\
Pre-VLA & \checkmark & -- & -- & -- & -- \\
Future--reality verif. & \checkmark & -- & $\sim$ & -- & -- \\
RTC & -- & -- & $\sim$ & \checkmark & -- \\
\midrule
\textbf{CheckVLA (ours)} & \checkmark & \checkmark & \checkmark & \checkmark & \checkmark \\
\bottomrule
\end{tabular*}
\caption{Execution-time capability comparison with SAFE \cite{safe}, Foresight \cite{foresight}, Pre-VLA \cite{pre_vla}, future--reality verification \cite{trust_imagination}, and RTC \cite{rtc}. \checkmark: core mechanism; $\sim$: partial (execution-length adaptation; switch-time continuity rather than deviation-triggered repair); --: absent. Marks reflect published mechanisms, not re-implementations.}
\label{tab:capability}
\vskip -0.1in
\end{table}

\section{Method}

\begin{figure*}[t]
\centering
\includegraphics[width=0.98\textwidth]{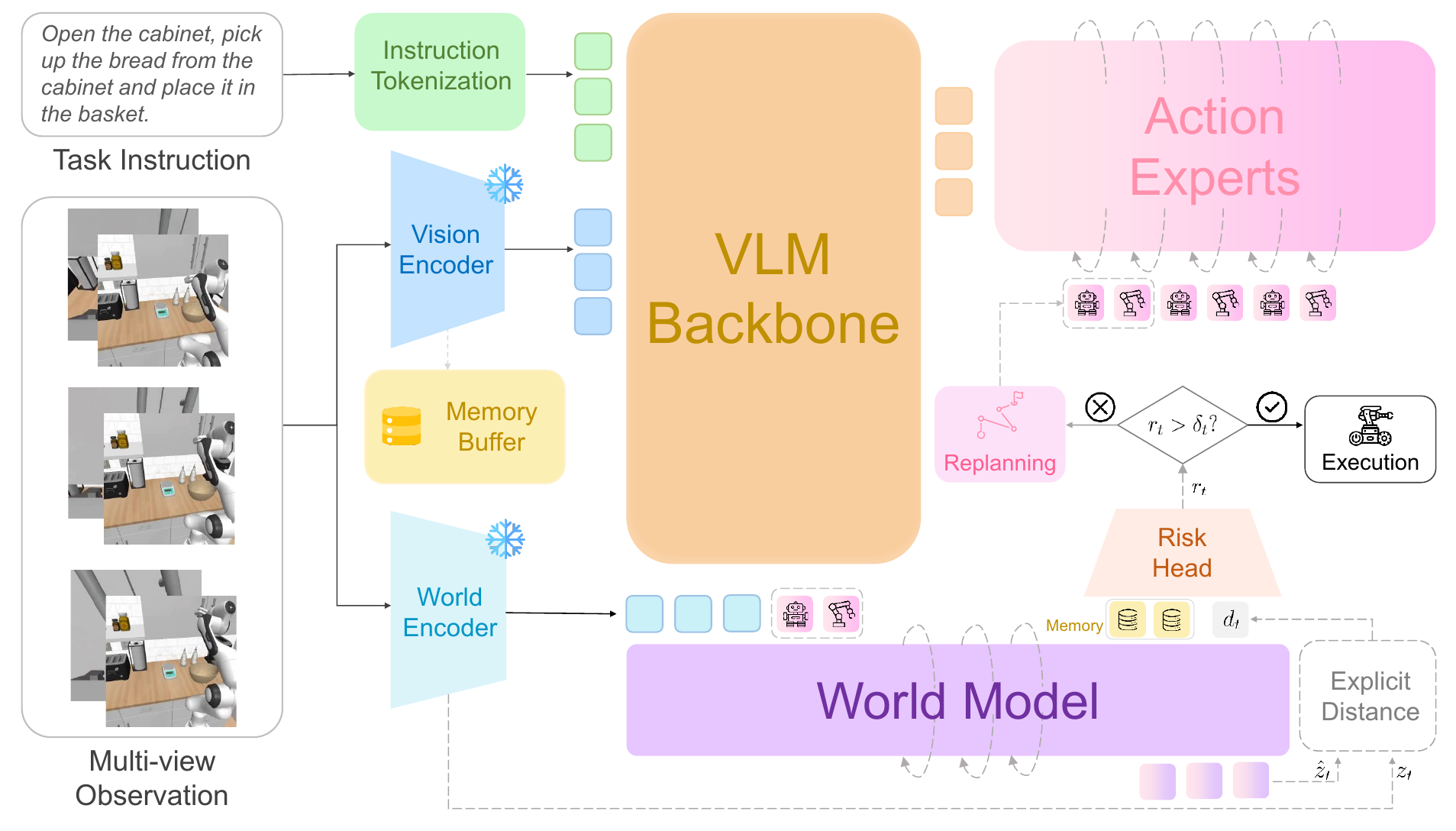}
\caption{The CheckVLA framework. A separately trained, frozen encoder supplies monitoring features; the world model predicts near-term consequences of the committed actions; the causal risk head compares aggregated risk $r_t$ with the calibrated threshold $\delta_t$; upon a declared failure, the same VLA rewrites the repairable suffix under the hard-prefix constraint with risk-graded guidance.}
\label{fig:method}
\vskip -0.15in
\end{figure*}

We introduce \textbf{CheckVLA}, a closed-loop execution framework that verifies the expected consequences of committed action chunks and repairs deviations while recovery remains feasible. As shown in Figure~\ref{fig:method}, four coupled components---action-conditioned rolling prediction, calibrated risk triggering, latency-aware suffix rewriting, and episodic context---implement the loop.

\subsection{Problem Formulation}
\label{sec:setup}

At a chunk boundary $t$, the policy $\pi_\theta$ encodes the instruction $L$, multi-view observation $o_t$, and episodic context $\rho_t$ as $c_t=\mathrm{VLM}(L,o_t;\rho_t)$ and commits
\begin{equation}
A_t=(a_{t,0},\dots,a_{t,H-1})\sim\pi_\theta\big(\cdot\mid c_t,\,s_t\big), \label{eq:chunk}
\end{equation}
where $s_t$ is proprioception and $A_t\in\mathbb{R}^{H\times M}$ spans mobility and manipulation. We use $h$ for a candidate position, $h_\tau$ for the active in-chunk index at control time $\tau$, and $\bar a_\tau$ for the action then committed for execution. The latest observation time $t'$ is an anchor, $\ell\le k$ is a prediction span, and $u\in[0,1]$ is flow-integration time. During a chunk, the high-level policy receives no new visual input \cite{act,diffusion_policy,pi0}, whereas the monitor does; repair is possible only while a useful suffix remains.

Suppose the first trigger occurs at $t^*$ with next old-chunk index $j=h_{t^*}$. The fixed schedule allocates $d_{\mathrm{lat}}$ control steps to inference, while $d_{\mathrm{eff}}$ denotes the elapsed steps when the replacement actually becomes deployable. Normally $d_{\mathrm{eff}}=d_{\mathrm{lat}}$; an early call waits, whereas an overrun makes $d_{\mathrm{eff}}>d_{\mathrm{lat}}$ and its elapsed replacement positions are discarded. Candidate positions $h<d_{\mathrm{eff}}$ are therefore irreversible, and only $h\ge d_{\mathrm{eff}}$ can be deployed (supplementary Secs.~\ref{sec:supp_execution_semantics}--\ref{sec:supp_pseudocode}).
\begin{figure}[t]
\centering
\includegraphics[width=\columnwidth]{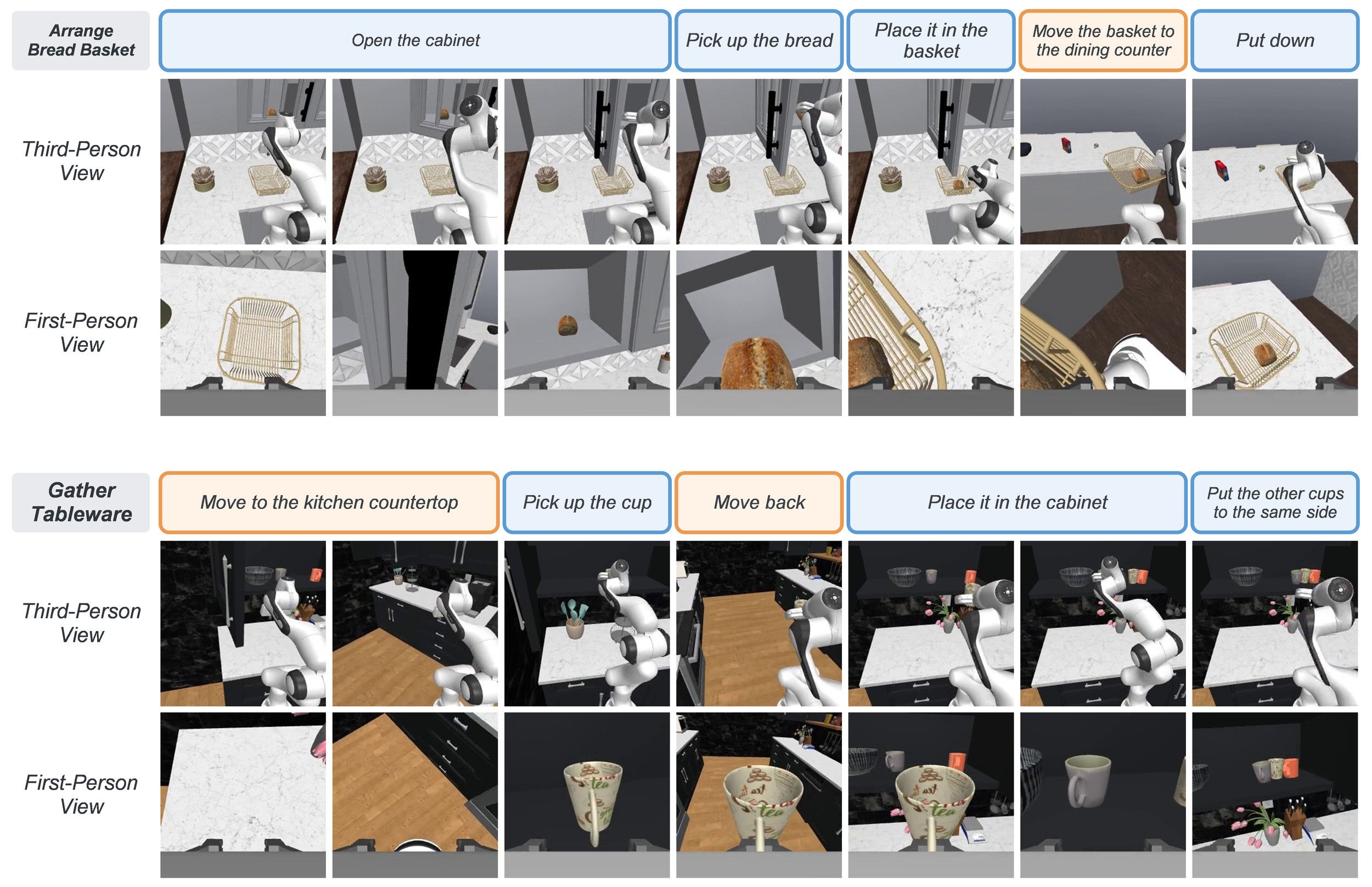}
\caption{Composite RoboCasa365 episodes (\textit{Arrange Bread Basket}, top; \textit{Gather Tableware}, bottom): subtask prompts aligned with third- and first-person keyframes; blue denotes manipulation and orange base motion.}
\label{fig:vis}
\vskip -0.2in
\end{figure}

\subsection{Action-Conditioned Rolling Prediction}
\label{sec:world_model}

While the chunk executes, each arriving observation restarts a short verification cycle: if the plan remains valid, future observations should agree with the action-conditioned predictions of the committed actions.

Let $z_{t'}=\phi(o_{t'})$ be the latest observation feature under a separately pretrained, frozen monitoring encoder $\phi$. Conditioned on the still-valid committed actions and nominal proprioceptive rollout $\hat s_{t'+i}=f(s_{t'},\bar a_{t':t'+i-1})$, the world model $\Psi$ predicts only the next $k$ steps ($k\ll H$):
\begin{equation}
\begin{aligned}
\hat z_{t'+i+1}=\Psi\big(&z_{t'},\hat z_{t'+1:t'+i},\\
&\bar a_{t':t'+i},\hat s_{t':t'+i}\big),\quad i=0,\ldots,k-1.
\end{aligned}\label{eq:rollout}
\end{equation}
Each observation re-anchors the rollout. Predictions crossing a chunk boundary are generated only after the next chunk is committed; after a rewrite, predictions conditioned on superseded actions are invalidated, and retrieval uses the latest valid anchor. Short rollouts limit autoregressive drift, while span-specific statistics below handle residual scale differences.

$\Psi$ is trained first by teacher forcing and then on its own rollouts, matching the online autoregressive context \cite{planet,dreamer,skyjepa}; training and prediction remain in feature space (supplementary Sec.~\ref{sec:supp_verifier}). The frozen encoder isolates monitoring from policy optimization. Although $\pi_\theta$ is stochastic, $\Psi$ provides a deterministic reference; empirical residual variation is absorbed by the fixed standardization and calibration pipeline.

\subsection{Calibrated Risk Triggering}
\label{sec:trigger}

As $z_t$ arrives, the monitor retrieves the prediction from the latest valid anchor with span $\ell(t)\le k$, computes $d_t=\mathrm{dist}\big(\hat z_{t\mid t-\ell(t)},z_t\big)$ on pooled, normalized features, and standardizes it with held-out-success discrepancy statistics:
\begin{equation}
\tilde d_t=\frac{d_t-\mu_d^{(\ell)}}{\max(\sigma_d^{(\ell)},\varepsilon)}. \label{eq:dist}
\end{equation}
The distance metric is selected on validation data; each observation then becomes a new anchor, and monitoring resumes at span one after a chunk switch.

A single-step discrepancy is weak evidence on its own, because contact, occlusion, and sensing noise all produce transient residuals. A temporally causal risk head $R$ therefore aggregates persistent and increasing patterns \cite{gauge} over the window $x_{t-w+1:t}$ of per-step tuples:
\begin{equation}
x_\tau=(\tilde d_\tau,\ \ell(\tau)/k,\ h_\tau/H,\ \Delta_{\mathrm{sw},\tau}),
\label{eq:tuples}
\end{equation}
where $h_\tau$ is the action index within the current chunk and $\Delta_{\mathrm{sw},\tau}$ is the number of control steps since the most recent chunk switch. The risk head additionally receives the episodic summary $m(\rho_t)$ introduced below, yielding
\begin{equation}
r_t=R\big(x_{t-w+1:t},\ m(\rho_t)\big)\in[0,1]. \label{eq:risk}
\end{equation}
The risk head is trained on nominally successful episodes, natural failures of the open-loop backbone, and onset-annotated physical perturbations; successes and benign deviations serve as negatives, and supervision is restricted to pre-failure evidence (objective and perturbation details in supplementary Sec.~\ref{sec:supp_verifier}).

Because one unnecessary intervention can spoil an otherwise successful episode, CheckVLA controls the probability of an unnecessary \emph{first} intervention over the episode. After fixing the policy, monitor, retrieval, discrepancy standardization, and memory rules, we fit risk statistics $(\mu_{r,t},\sigma_{r,t})$ on nominal-success shadow-mode trajectories disjoint from conformal calibration and set $\tilde\sigma_{r,t}=\max(\sigma_{r,t},\varepsilon)$. Split functional conformal calibration uses the per-trajectory scores $\sup_t(r_t^{(i)}-\mu_{r,t})/\tilde\sigma_{r,t}$ and their finite-sample quantile $\hat q_\alpha$ at miscoverage $\alpha$ \cite{foresight}. The online threshold is
\begin{equation}
\delta_t=\mu_{r,t}+\hat q_\alpha\,\tilde\sigma_{r,t}, \label{eq:threshold}
\end{equation}
and, for a new nominal-success trajectory exchangeable with calibration under this same fixed pipeline, satisfies the trajectory-marginal guarantee
\begin{equation}
\Pr\big(\exists\,t:\ r_t>\delta_t\mid\text{nominal success}\big)\le\alpha. \label{eq:guarantee}
\end{equation}
This is only a first-intervention guarantee: it covers neither failure recall, post-repair safety, repeated interventions, nor distribution shift. Those properties are evaluated empirically (supplementary Table~\ref{tab:reintervention}). The first crossing $t^*=\min\{t:r_t>\delta_t\}$ triggers repair.

\begin{table*}[t]
\centering
% \small
\setlength{\tabcolsep}{10pt}
\begin{tabular}{@{}lcccc@{}}
\toprule
Method & Atomic-Seen & Composite-Seen & Composite-Unseen & Average \\
\midrule
Diffusion Policy \cite{diffusion_policy}       & 15.7 & 0.2  & 1.3  & 6.1  \\
$\pi_0$ \cite{pi0}                            & 34.6 & 6.1  & 1.1  & 14.8 \\
$\pi_{0.5}$ \cite{pi05}                       & 39.6 & 7.1  & 1.2  & 16.9 \\
GigaWorld-Policy 0.1 \cite{gigaworld_policy}   & 44.4 & 11.8 & 2.9  & 20.7 \\
GR00T-N1.5 \cite{gr00t_n15}                    & 50.7 & 14.8 & 2.7  & 23.9 \\
GR00T-N1.6 \cite{gr00t_n16}                    & 51.1 & 9.4  & 1.7  & 21.9 \\
RLDX-1 \cite{rldx1}                            & 63.0 & \underline{27.5} & 5.4  & 33.2 \\
Qwen-RobotManip \cite{qwen_robotmanip}         & \textbf{68.6} & 20.1 & \textbf{14.9} & \underline{35.9} \\
WorldDreamer \cite{worlddreamer}               & \underline{66.3} & 26.7 & 9.0  & 35.3 \\
\midrule
\textbf{CheckVLA (ours)}                        & 63.7 & \textbf{30.9} & \underline{10.2} & \textbf{36.1} \\
\bottomrule
\end{tabular}
\caption{RoboCasa365 mean task success (\%) with policy pretraining limited to Human300, over 18 A-S, 16 C-S, and 16 C-U tasks. Average is task-count weighted; bold/underline denote the best/second-best. Published results provide descriptive positioning rather than paired comparison; CheckVLA additionally trains its verifier on training-side rollouts, and Table~\ref{tab:core_ablation} gives the controlled comparison.}
\label{tab:robocasa365_flagships}
\vskip -0.15in
\end{table*}

\subsection{Suffix Rewriting}
\label{sec:rewrite}

At $t^*$, execution continues while the same VLA generates a candidate indexed relative to the trigger. Its old-chunk reference is $A_{\mathrm{ref},h}=\bar a^{\mathrm{old}}_{t^*+h}$ for $0\le h<H-j$. At every flow-integration step, candidate positions $h<d_{\mathrm{lat}}$ are clamped to this executing prefix. Under nominal timing deployment starts at $h=d_{\mathrm{lat}}$; after an overrun it starts at $h=d_{\mathrm{eff}}$, discarding the extra elapsed positions. The hard clamp is an empirical projection, not exact conditional sampling (supplementary Sec.~\ref{sec:supp_rewrite}).

The retention strength of the superseded chunk is determined by the exceedance of the same calibrated threshold that produced the trigger. The remaining actions of that chunk provide a channel-wise reference, weighted by the \emph{standardized exceedance}
\begin{equation}
e_{t^*}=\Big(\big(r_{t^*}-\mu_{r,t^*}\big)\big/\tilde\sigma_{r,t^*}-\hat q_\alpha\Big)_{+}, \label{eq:exceedance}
\end{equation}
the amount by which the standardized risk exceeds the threshold; the mapping from exceedance to retention below is selected on validation data and carries no conformal guarantee. A base weight $w_0(e)=w_{\min}+(1-w_{\min})\exp(-\beta e)$, with $w_{\min}\in[0,1)$ and $\beta>0$, decreases monotonically in $e$: a small exceedance retains strong reference guidance, while a large one yields weaker retention and permits a larger correction. The weight decays with position at a channel-specific rate,
\begin{equation}
W_{h,m}=w_0(e_{t^*})\exp\big(-\lambda_m(h-d_{\mathrm{lat}})\big). \label{eq:guidance}
\end{equation}
This applies for $d_{\mathrm{lat}}\le h<H-j$ and $W=0$ otherwise, with rates $\lambda_m>0$ for the mobility and manipulation channels. Guidance mixes the reference into the flow velocity,
\begin{equation}
v=(1-W)\odot v_\theta+W\odot v_{\mathrm{ref}}, \label{eq:field}
\end{equation}
where $y$ is the current flow state, $\Delta u$ the integration step, and $v_{\mathrm{ref}}=(A_{\mathrm{ref}}-y)/\max(1-u,\Delta u)$. After switching, the loop resumes from the next observation (regular-transition guidance and suppression in supplementary Sec.~\ref{sec:supp_verifier}).

\subsection{Episodic Context}
\label{sec:memory}

A repair conditioned only on the current view may forget completed subgoals. CheckVLA therefore sets $\rho_t\equiv\mathcal{B}_t=\{(F_j,t_j)\}_{j=1}^{m_t}$, an event-driven bank of pooled policy features from real observations. Pause and diversity criteria govern writes across chunks and repairs. The policy reads the full bank through gated cross-attention, whereas the risk head receives only the compact summary $m(\rho_t)$; these policy features $F_j$ are distinct from the monitor features $z_t$ (supplementary Sec.~\ref{sec:supp_memory_impl}).

\subsection{Implementation}
\label{sec:implementation}

CheckVLA uses the $\pi_{0.5}$ flow-matching backbone \cite{pi05}. Mobility and manipulation experts have separate parameters but exchange information through joint self-attention over the shared VLM prefix \cite{abot_m05}; stop-gradient blocks their training signal from the VLM backbone \cite{knowledge_insulation}. Monitoring uses a frozen V-JEPA~2-AC encoder \cite{vjepa2}. Supplementary Secs.~\ref{sec:supp_action_experts}--\ref{sec:supp_memory_impl} describe the decoupled action experts, verifier training and calibration, and memory fusion.

\section{Experiments}

\subsection{Protocol and Public Benchmark}

We evaluate on RoboCasa365 \cite{robocasa365}, comprising 365 household mobile-manipulation tasks in 2,500 kitchens; Figure~\ref{fig:vis} shows two composite episodes. Following the official protocol, the action policy uses only Human300 (300 tasks, 100 human demonstrations each) and is tested on 18 Atomic-Seen (A-S), 16 Composite-Seen (C-S), and 16 Composite-Unseen (C-U) tasks. The verifier uses auxiliary nominal, natural-failure, and physics-perturbed rollouts from training-side tasks and scenes. Neither target-suite scenes nor C-U tasks train any component (splits and leakage guards in supplementary Sec.~\ref{sec:supp_protocol}).

Table~\ref{tab:robocasa365_flagships} lists published Human300 flagship results. CheckVLA achieves \textbf{36.1\%} average success, establishing a new state of the art among the methods considered: 19.2 percentage points above the published $\pi_{0.5}$ and 0.2 above Qwen-RobotManip. It ranks first on C-S and second on C-U while remaining competitive on A-S, consistent with execution verification benefiting tasks that chain subgoals. Because public systems differ beyond policy-pretraining data, this state-of-the-art positioning is descriptive; the controlled attribution follows.

\begin{table*}[t]
\centering
% \small
\setlength{\tabcolsep}{7pt}
\begin{tabular}{@{}lcccccc@{}}
\toprule
Variant & A-S & C-S & C-U & Average & Calls/ep. & $\Delta$ Avg. \\
\midrule
Reproduced $\pi_{0.5}$                         &39.8 &7.4  &1.5  &17.2 &7.8  & -- \\
Capacity-matched monolithic expert             &45.9 &10.8 &2.8  &20.9 &7.8  &+3.7 \\
Decoupled action backbone, open loop            &46.8 &11.9 &3.1  &21.6 &7.8  &+0.7 \\
$+$ regular-transition guidance                 &48.7 &13.6 &3.8  &23.1 &7.8  &+1.5 \\
$+$ invocation-matched periodic replan          &53.9 &19.8 &5.8  &27.6 &10.1 &+4.5 \\
$+$ verified trigger, fixed suffix guidance     &58.8 &24.7 &7.5  &31.5 &10.0 &+3.9 \\
$+$ risk-adaptive guidance, no memory            &61.5 &28.3 &9.0  &34.1 &10.0 &+2.6 \\
\textbf{Full CheckVLA}                          & \textbf{63.7} & \textbf{30.9} & \textbf{10.2} & \textbf{36.1} &10.2 &+2.0 \\
\bottomrule
\end{tabular}
\caption{Sequential build-up on RoboCasa365 under a common training recipe and paired seeds. Average is task-count weighted; Calls/ep. reports the invocation budget. Supplementary Secs.~\ref{sec:supp_detector}--\ref{sec:supp_failures} provide detector, binding, repair, robustness, memory, runtime, and failure audits.}
\label{tab:core_ablation}
\vskip -0.2in
\end{table*}

\begin{figure*}[t]
\centering
\includegraphics[width=0.98\textwidth]{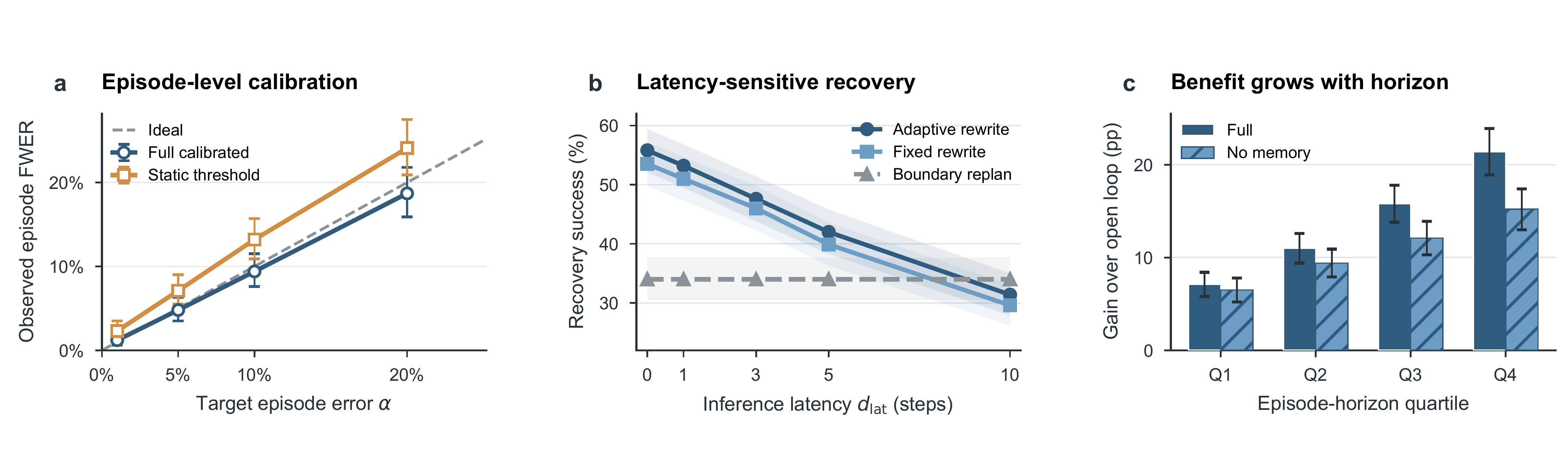}
\vskip -0.1in
\caption{Closed-loop diagnostics. (a) Episode-level unnecessary-first-intervention frequency versus target conformal level. (b) Perturbation recovery across inference latency. (c) Paired gain over open loop across horizon quartiles, with and without memory. Bars and bands are 95\% intervals: exact binomial in (a), hierarchical paired bootstrap in (b) and (c).}

\label{fig:diagnostics}
\vskip -0.15in
\end{figure*}

\subsection{Fair Internal Controls}

Table~\ref{tab:core_ablation} builds the system under common splits, optimization, and paired evaluation seeds, first controlling action-model capacity and then adding transition guidance, periodic replanning, verified timing, adaptive guidance, and memory. Changed policy inputs are retrained and affected monitors recalibrated. Validation selects the periodic interval, yielding 10.1 calls per episode versus 10.2 for CheckVLA. Replacing periodic timing with the verified trigger under fixed guidance adds +3.9 points; risk-adaptive guidance and memory add +2.6 and +2.0.

\begin{table}[t]
\centering
\small
\setlength{\tabcolsep}{1.2pt}
\begin{tabular*}{\columnwidth}{@{\extracolsep{\fill}}lccc@{}}
\toprule
Detector
& \shortstack{Episode\\FWER (\%)}
& \shortstack{Timely\\recall (\%) $\uparrow$}
& \shortstack{Perturbed\\success (\%) $\uparrow$} \\
\midrule
MC action entropy & 5.0 & 58.7 & 38.9 \\
Flow-path variance & 5.2 & 61.4 & 40.2 \\
Sampling disagreement & 4.9 & 64.8 & 41.5 \\
Policy-feature probe & 5.1 & 68.6 & 43.7 \\
\midrule
Obs.-only world predictor & 5.3 & 48.6 & 35.2 \\
Action-shuffled predictor & 5.1 & 37.9 & 31.0 \\
\midrule
No rolling re-anchor & 5.4 & 62.8 & 38.7 \\
Instantaneous score & 5.0 & 66.1 & 41.2 \\
Constant threshold & 7.1 & 71.5 & 45.0 \\
\midrule
\textbf{Full action-cond.\ verifier} & \textbf{4.8} & \textbf{77.9} & \textbf{47.6} \\
\bottomrule
\end{tabular*}
\caption{Trigger comparison at target episode FWER $\alpha=0.05$ with a shared rewriter and latency. Policy-side scores are evaluated at commit time; the constant threshold is validation-tuned, not conformally calibrated; the third block ablates verifier mechanisms.}
\label{tab:trigger_comparison}
\vskip  -0.2in
\end{table}

\subsection{Ablation Studies and Deployment Audits}

A controlled study injects physical perturbations across task phases in 12 composite tasks. Conformal detector variants are independently recalibrated at the common target episode FWER, detector comparisons replay identical shadow trajectories, and rewrite variants branch from the same first-trigger snapshot with shared randomness. Episode FWER is the fraction of nominal successes with an unnecessary first intervention. A detection is timely when it crosses after annotated onset and deploys before the end of the empirically estimated recovery window under the tested repair set. Rescue and harm are paired outcome changes from the trigger snapshot (supplementary Sec.~\ref{sec:supp_metrics}); the constant-threshold control is deliberately nonconformal.

\paragraph{Action conditioning.} To separate action--consequence evidence from observation novelty, we retrain an action-shuffled world model that preserves action-token marginals and an observation-only predictor. Timely recall falls from 77.9\% to 37.9\% and 48.6\%, and perturbed success from 47.6\% to 31.0\% and 35.2\% (Table~\ref{tab:trigger_comparison}). These controls support the joint action-conditioned pathway; a frozen-model counterfactual audit confirms action--future binding without separating action tokens from their deterministic proprioceptive rollout (supplementary Sec.~\ref{sec:supp_action_binding}).

\paragraph{Verifier mechanisms.} Removing rolling re-anchoring or temporal aggregation costs 15.1 and 11.8 points of timely recall, respectively; a validation-tuned constant threshold raises FWER from 4.8\% to 7.1\% (Table~\ref{tab:trigger_comparison}; Figure~\ref{fig:diagnostics}a). Together these ablations support the composite design: short re-anchored spans and persistence aggregation improve timeliness, while functional calibration sets the deployed operating point.

\paragraph{Policy-side alternatives.} We calibrate four commit-time detectors under the same protocol: Monte Carlo action entropy, flow-path variance, sampling disagreement, and a frozen policy-feature probe. The full verifier exceeds the strongest by 9.3 points in timely recall and 3.9 in perturbed success; refreshing policy scores at the observation rate would add 382.2 VLA evaluations per episode (supplementary Sec.~\ref{sec:supp_detector}). Thus post-dispatch action-conditioned evidence adds information at monitor rather than policy cost.

\paragraph{Confidently-wrong quadrant.} Table~\ref{tab:confidently_wrong} splits held-out natural executions by independently calibrated commit-time uncertainty and world-model risk. The low-uncertainty, high-risk quadrant contains 25.0\% of episodes but 48.4\% of failures; 76.0\% remain timely recoverable and 31.0\% are rescued by the fixed rule. This operational pattern supports complementarity between the signals. Natural-failure and benign-shift tests probe distribution sensitivity, while leave-one-family-out, severity, and data-scale studies probe monitor generalization (supplementary Secs.~\ref{sec:supp_external_validity} and~\ref{sec:supp_sensitivity}).

\begin{table}[t]
\centering
\small
\setlength{\tabcolsep}{1.5pt}
\begin{tabular*}{\columnwidth}{@{\extracolsep{\fill}}llccccc@{}}
\toprule
\shortstack{Policy\\unc.} & \shortstack{WM\\risk}
& \shortstack{Ep.\\share (\%)}
& \shortstack{Fail.\\rate (\%)}
& \shortstack{Fail.\\share (\%)}
& \shortstack{Timely\\recov. (\%)}
& \shortstack{Rescue\\(\%) $\uparrow$} \\
\midrule
Low  & Low  & 52.0 & 12.0 & 25.2 & 18.0 & 6.0 \\
High & Low  & 14.0 & 18.0 & 10.2 & 28.0 & 11.0 \\
\textbf{Low} & \textbf{High} & \textbf{25.0} & \textbf{48.0} & \textbf{48.4} & \textbf{76.0} & \textbf{31.0} \\
High & High & 9.0 & 45.0 & 16.2 & 73.0 & 29.0 \\
\bottomrule
\end{tabular*}
\caption{Commit-time policy uncertainty versus world-model risk on held-out natural executions assignable at the preregistered decision window (supplementary Sec.~\ref{sec:supp_external_validity}), each signal split at its independently calibrated threshold; the low-uncertainty, high-risk cell is the confidently-wrong quadrant.}
\label{tab:confidently_wrong}
\vskip -0.1in
\end{table}

\paragraph{Repair rules.} Six rewrite rules and a wait-for-boundary baseline branch from identical first-trigger snapshots: unconstrained generation with post-hoc prefix disposal, hard-prefix full rewrite, RTC inpainting~\cite{rtc}, fixed guidance, adaptive guidance with shuffled exceedance, and correctly paired adaptive guidance (Table~\ref{tab:repair_main}). The paired rule reaches 16.9\% rescue at 2.8\% harm, versus 12.8\%/3.7\% for fixed guidance and 14.4\%/3.6\% when exceedance is shuffled (Figure~\ref{fig:diagnostics}b; supplementary Secs.~\ref{sec:supp_rewrite}--\ref{sec:supp_exceedance_bins}). The pairing, not merely variable weights, improves both outcomes.

\paragraph{Episodic memory.} A fully retrained and, where needed, recalibrated policy-bank $\times$ risk-summary factorial avoids the confounding of test-time zeroing. The full design cuts completed-subgoal regressions from 1.34 to 0.18 per episode, with success gains increasing over the horizon (Figure~\ref{fig:diagnostics}c). The policy reader drives most progress retention; the smaller risk-summary effect brings empirical FWER closer to target.

\paragraph{Deployment cost.} The full system adds 88.4M monitor parameters and 16.9 GFLOPs per step, runs at 1.18$\times$ the open-loop wall-clock cost, and keeps monitor p95 latency below one control period. It intervenes in 4.8\% of clean episodes and harms 2.1\% of otherwise successful clean continuations; on held-out natural executions, success rises from 61.4\% to 68.9\% at 2.7\% harm (supplementary Tables~\ref{tab:natural_failure}, \ref{tab:reintervention}, and~\ref{tab:runtime}). A blinded audit assigns residual failures to detection, physics, policy, or repair causes (supplementary Sec.~\ref{sec:supp_failures}).
\begin{table}[t]
\centering
\small
\setlength{\tabcolsep}{2pt}
\begin{tabular*}{\columnwidth}{@{\extracolsep{\fill}}lcccc@{}}
\toprule
Rewrite rule
& \shortstack{Rescue\\(\%) $\uparrow$}
& \shortstack{Harm\\(\%) $\downarrow$}
& \shortstack{Switch\\jump $\downarrow$}
& \shortstack{Success\\(\%) $\uparrow$} \\
\midrule
Wait for boundary & 7.4 & 4.8 & -- & 34.0 \\
Unconstrained, discard prefix & 10.1 & 8.9 & 0.284 & 39.8 \\
Hard prefix, full rewrite & 13.2 & 6.8 & 0.171 & 42.4 \\
RTC inpainting & 13.7 & 4.1 & 0.091 & 45.6 \\
Fixed guidance $W$ & 12.8 & 3.7 & 0.104 & 46.0 \\
Shuffled exceedance & 14.4 & 3.6 & 0.103 & 44.3 \\
\midrule
\textbf{Adaptive $W(e)$} & \textbf{16.9} & \textbf{2.8} & \textbf{0.072} & \textbf{47.6} \\
\bottomrule
\end{tabular*}
\caption{Fixed-trigger repair at nominal latency ($d_{\mathrm{lat}}{=}3$), branched from identical first-trigger snapshots. Rescue, harm, and switch jump are one-shot outcomes with subsequent triggers disabled; success continues the closed loop (full latency sweep in supplementary Sec.~\ref{sec:supp_rewrite}).}
\label{tab:repair_main}
\vskip -0.15in
\end{table}
\section{Discussion}

Our findings suggest that an action chunk is both a control command and a testable prediction of future observations. A separately trained action-conditioned verifier can restore feedback during open-loop execution by deciding when another policy call is warranted. Threshold exceedance then determines how strongly to revise the committed suffix. This is not a general safety guarantee, because functional conformal calibration controls only the probability of an unnecessary first intervention on exchangeable nominal-success episodes. It does not guarantee recall, post-repair safety, or coverage under distribution shift. Repair is also constrained by latency and policy capability: if too little deployable suffix remains or no viable replacement exists, even a correct warning cannot recover the episode. Evidence is currently limited to RoboCasa365 simulation, a finite perturbation family, and a frozen representation. Changes in sensing, hardware timing, or task distribution require recalibration and retesting. Dynamic latency-aware scheduling and hardware validation are therefore the natural next steps.

\section{Conclusion}

We propose CheckVLA, which reframes a committed action chunk as a testable prediction and verifies it during execution with a separately trained action-conditioned world model, restoring feedback during chunked execution. At a matched policy-call budget on RoboCasa365, it improves average success by 8.5 percentage points over periodic replanning; a fixed-guidance ablation attributes 3.9 points to verified timing. These results suggest that execution-time verification complements policy scaling by restoring feedback that a stronger open-loop policy alone does not provide.

% References are inlined so this source has no BibTeX database dependency.
\FloatBarrier

% The complete technical appendix is inlined below.
\clearpage
\appendix
% Appendix setup for this standalone arXiv manuscript.
% Keep wide floats close to their discussion without letting several of them
% consume an otherwise text-bearing page.
\renewcommand{\topfraction}{0.88}
\renewcommand{\dbltopfraction}{0.72}
\renewcommand{\bottomfraction}{0.75}
\renewcommand{\textfraction}{0.15}
\renewcommand{\floatpagefraction}{0.8}
\renewcommand{\dblfloatpagefraction}{0.85}
\setcounter{topnumber}{2}
\setcounter{dbltopnumber}{2}
\setcounter{bottomnumber}{2}
\setcounter{totalnumber}{4}
\setlength{\textfloatsep}{13pt plus 2pt minus 2pt}
\setlength{\floatsep}{12pt plus 2pt minus 2pt}
\setlength{\dbltextfloatsep}{13pt plus 2pt minus 2pt}
\setlength{\dblfloatsep}{12pt plus 2pt minus 2pt}
% Float-only pages, when unavoidable, should pack from the top rather than
% stretching several tables across the full page height.
\makeatletter
\setlength{\@fptop}{0pt}
\setlength{\@fpsep}{12pt}
\setlength{\@fpbot}{0pt plus 1fil}
\setlength{\@dblfptop}{0pt}
\setlength{\@dblfpsep}{12pt}
\setlength{\@dblfpbot}{0pt plus 1fil}
\makeatother
\setcounter{secnumdepth}{1}
\renewcommand{\thesection}{\Alph{section}}
\setcounter{figure}{0}
\setcounter{table}{0}
\setcounter{equation}{0}
\setcounter{algorithm}{0}
\renewcommand{\thefigure}{S\arabic{figure}}
\renewcommand{\thetable}{S\arabic{table}}
\renewcommand{\theequation}{S\arabic{equation}}
\renewcommand{\thealgorithm}{S\arabic{algorithm}}

\section*{Appendix}

\begin{figure*}[!tbp]
\centering
\includegraphics[width=0.92\textwidth]{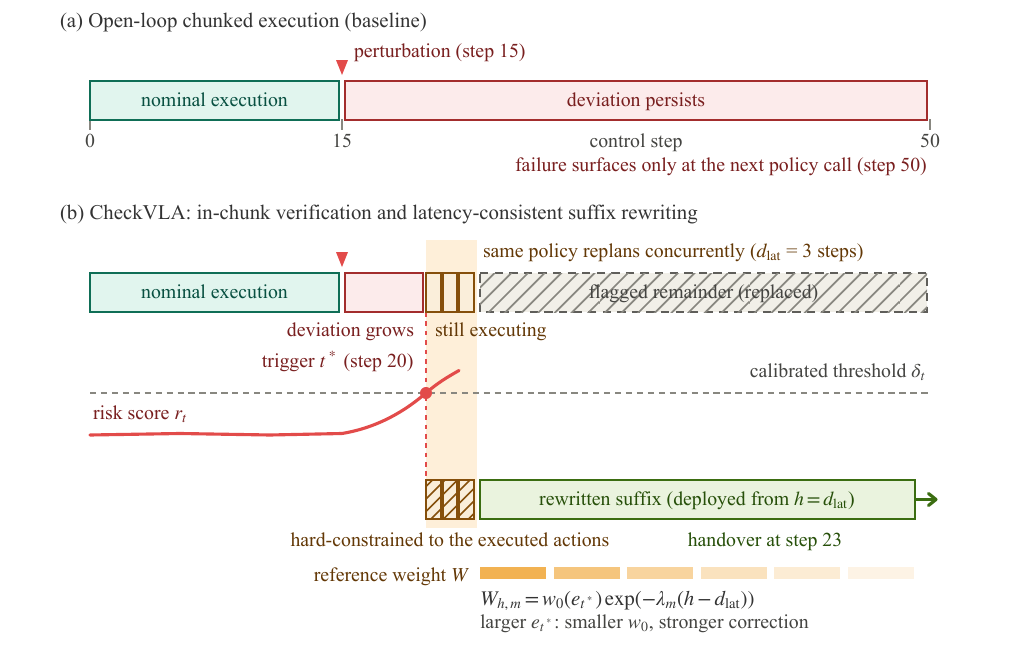}
\caption{Suffix rewriting under inference latency. (a)~Open-loop chunked execution: a perturbation at step~15 is not observed at the policy level, and the deviation persists until the next policy call at the chunk boundary; the task fails. (b)~CheckVLA: the action-conditioned risk score $r_t$ crosses the calibrated threshold $\delta_t$ at $t^*$ (step~20) and triggers replanning by the same policy. While inference runs ($d_{\mathrm{lat}}{=}3$ control steps), the flagged chunk keeps executing (irreversible prefix); the new chunk is hard-constrained to match these executed actions and deploys from $h{=}d_{\mathrm{lat}}$ (step~23), replacing the flagged remainder. The guidance weight starts at $w_0(e_{t^*}){<}1$ and decays along the suffix: a larger standardized exceedance permits a stronger correction.}
\label{fig:suffix_timeline}
\end{figure*}

\section{Execution-Semantics Illustration}
\label{sec:supp_execution_semantics}

Figure~\ref{fig:suffix_timeline} gives a schematic episode that illustrates the execution semantics formalized around Eq.~\ref{eq:chunk} and the suffix-rewriting mechanism in Eqs.~\ref{eq:guidance}--\ref{eq:field}. Both panels share the same horizontal axis (control steps 0--50, one chunk length $H$) and receive the same perturbation: at step 15, the carried object slips inside the gripper. Panel (a) shows open-loop chunked execution: the policy receives no new high-level visual input within the chunk, the deviation accumulates from step 15 onward, and the failure can surface only at the next policy call at step 50. Panel (b) shows the same episode under CheckVLA. The world model predicts near-term features conditioned on the committed actions; the prediction--observation discrepancies are aggregated by the temporally causal risk head into the risk score $r_t$ (red curve), which rises after the perturbation and crosses the conformally calibrated, time-varying threshold $\delta_t$ (gray dashed line) at $t^*$ (step 20), triggering a replan. A trigger does not mean immediate correction: one inference of the same policy takes $d_{\mathrm{lat}}=3$ control steps (amber band), during which old-chunk steps 20--22 still execute and form the irreversible prefix. Here $d_{\mathrm{lat}}$ is a fixed deployment schedule chosen to cover the median measured latency (286\,ms at the 100\,ms control period; Table~\ref{tab:runtime}): the switch is scheduled $d_{\mathrm{lat}}$ ticks after the trigger, an earlier-finishing call waits for the scheduled tick, and a call that overruns deploys at the first tick after completion with the elapsed replacement positions discarded (the effective delay $d_{\mathrm{eff}}$ defined immediately after Eq.~\ref{eq:chunk}), so latency jitter only shortens the deployed suffix. The reported experiments impose the nominal schedule uniformly, and the sweep in Table~\ref{tab:rewrite_latency} brackets slower calls up to $d_{\mathrm{lat}}{=}10$. The corresponding prefix positions of the new chunk are hard-constrained to these executed actions (hatched cells), reducing the command discontinuity at the switch; the new chunk deploys from $h=d_{\mathrm{lat}}$ (step 23, green), and the remainder of the old chunk (gray hatching) is replaced and never executes. The bottom decay strip shows the per-position guidance weight $W$ on the flagged old chunk: its entry scale $w_0(e_{t^*})<1$ is set by the standardized exceedance at the trigger and decays as $\exp(-\lambda_m(h-d_{\mathrm{lat}}))$; the larger the exceedance, the smaller $w_0$ and the stronger the permitted correction. The first deployed position therefore departs from the flagged reference instead of copying it. Step indices are illustrative values from the experimental configuration ($H{=}50$, $d_{\mathrm{lat}}{=}3$); the real distributions of perturbation types, trigger times, and lead times are reported in supplementary Sec.~\ref{sec:supp_protocol}.

\section{Execution-Loop Pseudocode}
\label{sec:supp_pseudocode}

Algorithm~\ref{alg:checkvla} consolidates the monitoring, triggering, suppression, and re-anchoring rules formalized across Eqs.~\ref{eq:chunk}--\ref{eq:field} into a single loop; every equation reference below resolves to a labeled equation in the main text.

\begin{algorithm}[!htbp]
\caption{CheckVLA execution loop (one episode)}
\label{alg:checkvla}
\begin{algorithmic}[1]
\REQUIRE policy $\pi_\theta$, frozen encoder $\phi$, world model $\Psi$, risk head $R$, calibrated $(\mu_{r,t},\tilde\sigma_{r,t},\hat q_\alpha)$, span $k$, scheduled latency $d_{\mathrm{lat}}$, suppression length $s_{\min}$
\STATE commit $A_t\sim\pi_\theta(\cdot\mid c_t,s_t)$ (main Eq.~\ref{eq:chunk}); anchor at the current observation
\WHILE{episode not finished}
\STATE \textbf{on} each new observation $o_t$: $z_t\leftarrow\phi(o_t)$; write the corresponding pooled policy feature $F_t$ to the keyframe bank $\rho_t$ if the pause and diversity criteria hold
\STATE retrieve the prediction $\hat z_{t\mid t-\ell(t)}$ issued at the latest valid anchor; standardize the discrepancy by span (main Eq.~\ref{eq:dist}) and form the tuple $x_t$ (main Eq.~\ref{eq:tuples})
\STATE $r_t\leftarrow R\big(x_{t-w+1:t},\,m(\rho_t)\big)$;\quad $\delta_t\leftarrow\mu_{r,t}+\hat q_\alpha\,\tilde\sigma_{r,t}$ (main Eqs.~\ref{eq:risk}--\ref{eq:threshold})
\IF{$r_t>\delta_t$, no rewrite in progress, and no post-deployment suppression}
\STATE compute the standardized exceedance $e_{t^*}$ (main Eq.~\ref{eq:exceedance}); rewrite with $\pi_\theta$: clamp positions $h<d_{\mathrm{lat}}$ to the executing prefix at every integration step and guide the suffix with $W_{h,m}$ (main Eqs.~\ref{eq:guidance}--\ref{eq:field})
\STATE at the scheduled tick deploy from $h=d_{\mathrm{lat}}$; if inference overruns, set $d_{\mathrm{eff}}$ to elapsed ticks, deploy from $h=d_{\mathrm{eff}}$, and discard earlier positions; invalidate superseded predictions, suppress triggers for $s_{\min}$ steps, and re-anchor at the next observation
\ELSE
\STATE anchor at $z_t$; roll out $\hat z$ for the next $k$ steps conditioned on the remaining committed actions (main Eq.~\ref{eq:rollout})
\ENDIF
\STATE \textbf{at} a regular chunk boundary: commit the next chunk with bank $\rho_t$ ($e{=}0$ transition guidance)
\ENDWHILE
\end{algorithmic}
\end{algorithm}

\section{Decoupled Action Experts and Training Details}
\label{sec:supp_action_experts}

\begin{figure}[!htbp]
\centering
\includegraphics[width=0.94\columnwidth]{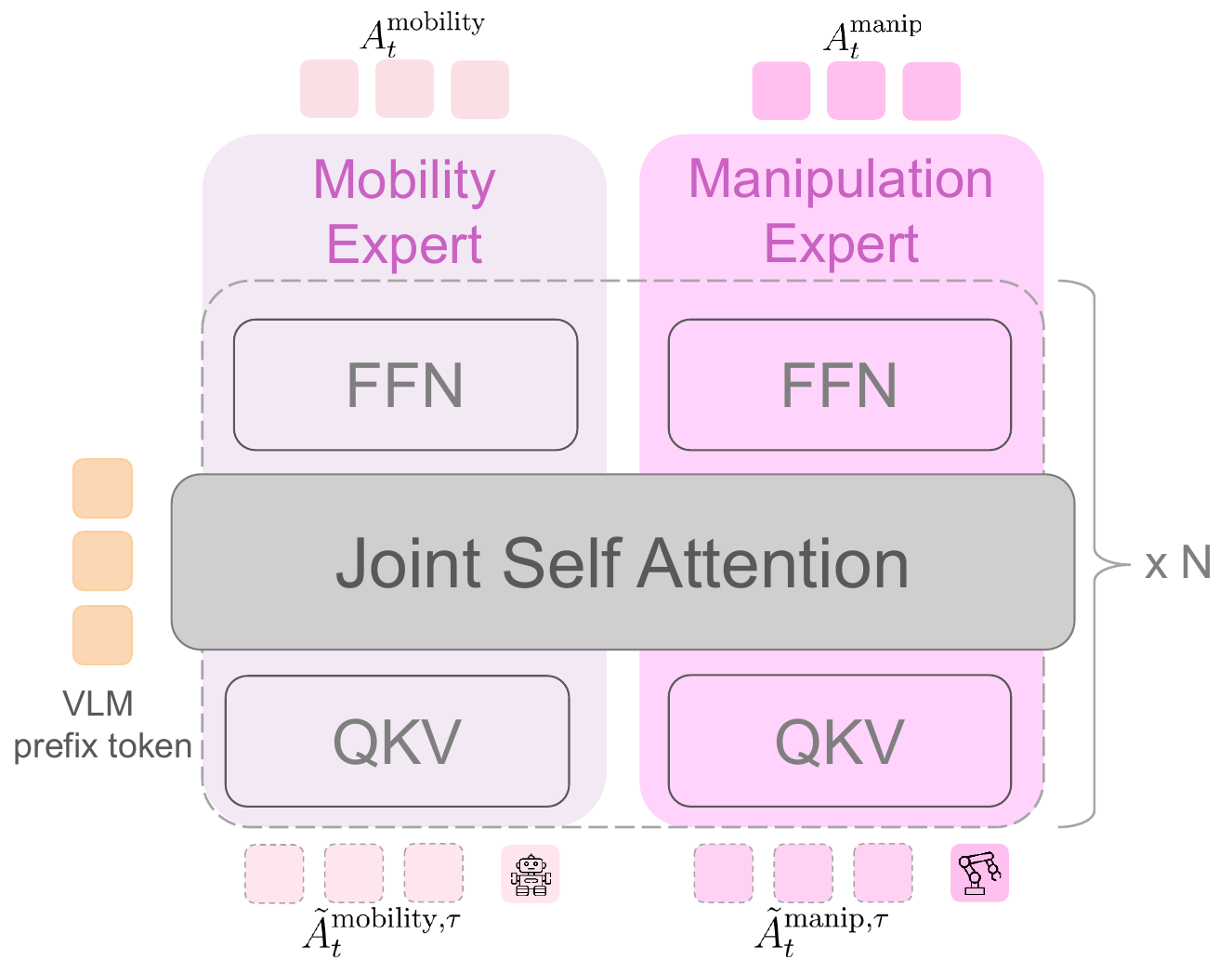}
\caption{Decoupled action experts. The mobility and manipulation experts keep separate QKV projections, feed-forward networks, and denoising heads, and interact only through joint self-attention over the shared VLM prefix tokens, repeated over $N$ blocks. Each expert denoises its own noisy action tokens $\tilde A_t^{e,u}$ (Eq.~\ref{eq:interp}) into the committed subspace actions $A_t^{e}$.}
\label{fig:action_experts}
\end{figure}

Action generation uses decoupled mobility and manipulation experts (Figure~\ref{fig:action_experts}). The mobility expert receives base-state and base-action tokens, while the manipulation expert receives arm--gripper state and action tokens; each expert has separate query/key/value projections, a feed-forward network, and a denoising head. Their projected tokens interact in joint self-attention over the shared prefix $c_t$, allowing coordination without additional cross-expert connections; the design follows evidence that action-subspace decoupling can reduce gradient interference~\cite{abot_m05}.

 For $e\in\{\text{mobility},\text{manip}\}$, we sample a flow-integration time $u\sim\mathcal{U}(0,1)$ and form noisy actions by linear interpolation:
\begin{equation}
\tilde A_t^{e,u}=u A_t^{e}+(1-u)\,\epsilon^{e},\qquad \epsilon^{e}\sim\mathcal{N}(0,I), \label{eq:interp}
\end{equation}
and training minimizes the conditional flow-matching loss
\begin{equation}
\mathcal{L}_{\mathrm{AE}}=\sum_{e}\mathbb{E}_{u,\epsilon^{e}}\Big\|v_\theta^{e}\big(\tilde A_t^{e,u},s_t^{e};\,c_t,u\big)-\big(A_t^{e}-\epsilon^{e}\big)\Big\|^2, \label{eq:fm}
\end{equation}
where $s_t^{e}$ is the corresponding proprioceptive state and $u$ is injected through adaptive layer normalization. A stop-gradient isolates the VLM backbone from the flow-matching objective. At inference, the experts synchronously integrate their vector fields under the shared prefix; a replan trigger invokes the same network with the guided flow integration in Eqs.~\ref{eq:guidance}--\ref{eq:field}.

\section{Verifier Training and Calibration Details}
\label{sec:supp_verifier}

The world model $\Psi$ applies block-causal attention over the interleaved feature and action tokens, with block $i$ predicting frame $i{+}1$. Training first minimizes a teacher-forcing objective with a Huber loss against frozen encoder targets,
\begin{equation}
\mathcal{L}_{\Psi}^{\mathrm{tf}}=\sum_{i=1}^{k}\ell_{\mathrm{Huber}}\big(\hat z_{t+i},\ \mathrm{sg}(z_{t+i})\big), \label{eq:supp_tf}
\end{equation}
and then replaces the prediction context with $\Psi$'s own rollouts so that training matches the autoregressive context used online. Predictions are stored by anchor and target time, and targets that cross a chunk boundary are generated after the next chunk is committed; no pixel decoding is used at any stage. At inference, the monitor retrieves the prediction from the most recent anchor whose rollout is available and whose conditioning actions have not been superseded. Under nominal timing (monitor p95 of 16.4\,ms against the 100\,ms control period; Table~\ref{tab:runtime}) this is the previous observation, so the retrieved span is $\ell=1$; spans up to $k$ arise when monitor computation overruns a control period, an observation drops, or a chunk switch invalidates predictions, and the span-specific statistics of main Eq.~\ref{eq:dist} keep these regimes on a common scale.

The risk head is trained on three trajectory classes: nominally successful episodes, natural failures of the open-loop backbone, and physics-level perturbations---impulses to objects or the base, transient joint offsets, and object displacements---annotated by their onset $\tau_{\mathrm{on}}$. The objective applies binary cross-entropy to the softmax-pooled risk within $[\tau_{\mathrm{on}},\,\tau_{\mathrm{on}}{+}\Delta]$, penalizes risk before onset, and masks gradients after task failure; nominally successful episodes and benign deviations (after which the task still succeeds) serve as negatives throughout. Supervision is therefore restricted to pre-failure evidence, and perturbations act at the physics level; they are never injected directly into the discrepancy sequence.

The functional conformal threshold in Eqs.~\ref{eq:threshold}--\ref{eq:guarantee} is constructed as follows. Per-step risk statistics $(\mu_{r,t},\sigma_{r,t})$ are fitted on nominally successful shadow-mode trajectories of the same training seed, disjoint from the $n$ calibration trajectories, with $\tilde\sigma_{r,t}=\max(\sigma_{r,t},\varepsilon)$. Each calibration trajectory $i$ receives the score $\kappa^{(i)}=\sup_t\big(r_t^{(i)}-\mu_{r,t}\big)/\tilde\sigma_{r,t}$, and $\hat q_\alpha$ is the $\lceil(n{+}1)(1{-}\alpha)\rceil$-th order statistic of these scores, set to $+\infty$ when the index exceeds $n$. Every score-shaping statistic, including discrepancy-span statistics and temporal normalization, and every online retrieval and memory rule is fixed before the conformal quantile is computed; exchangeability between calibration and deployment episodes under the same fixed pipeline then yields the first-intervention episode-level bound in Eq.~\ref{eq:guarantee}.

During regular chunk transitions, reference guidance uses $e=0$, hence $w_0=1$, with a slower positional decay to prioritize continuity. After a repair switch, predictions conditioned on superseded actions are invalidated and the next observation forms a new anchor; further triggers are suppressed while a rewrite is in progress and for $s_{\min}$ steps after deployment, and the reference weight $W$ is zero beyond the end of the remaining reference actions. Table~\ref{tab:config} lists the complete hyperparameter configuration.

\section{Episodic Context Implementation}
\label{sec:supp_memory_impl}

A post-replan invocation sees only the current observation, while evidence of completed stages may have left the field of view. CheckVLA therefore maintains an event-driven keyframe bank $\mathcal{B}_t=\{(F_j,t_j)\}_{j=1}^{m_t}$ of real-observation features that persists across chunks and repairs and is cleared only at episode end. A non-learned two-stage filter proposes writes: delay-confirmed local minima of the mean joint-space displacement $\overline{\Delta q}_t=\frac{1}{w_s}\sum_{j=t-w_s+1}^{t}\lVert q_j-q_{j-1}\rVert_2$ mark candidate pauses, and candidates too similar or too close in time to the latest stored frame are rejected. Accepted entries store pooled policy-encoded tokens $F_j$ with timestamps; positional indices are rebuilt from bank order at read time.

The policy needs the evidence itself to choose an action, whereas the risk head needs only enough context to judge whether a discrepancy is innocuous. The policy therefore fuses the full bank once per invocation through gated cross-attention,
\begin{equation}
\begin{aligned}
X_t'&=\mathrm{CrossAttn}\big(X_t,\ \tilde{\mathcal{B}}_t\big),\\
\hat X_t&=X_t+\sigma\big(g([X_t;X_t'])\big)\odot X_t',
\end{aligned} \label{eq:supp_fusion}
\end{equation}
where $g$ is negatively initialized and $\hat X_t$ preserves the token shape expected by the VLM. The risk head reads only $m(\rho_t)=\big(t-t_{\mathrm{last}},\ m_t,\ \bar F_{\mathrm{last}}\big)$. Both paths are trained with real banks reconstructed offline by the same filter. Consequently, the factorial memory study retrains each policy-reader variant and independently recalibrates each changed risk reader; test-time zeroing is not treated as a valid substitute.

\section{Evaluation Protocol and Data Separation}
\label{sec:supp_protocol}

The supplementary evaluation is designed to test one claim: under the same action backbone and comparable intervention budgets, action-conditioned verification should identify useful intervention times, and latency-consistent suffix rewriting should turn those warnings into recoveries without increasing harm on episodes that would otherwise succeed. This claim is bounded to the RoboCasa365 simulator and to the task, scene, and perturbation distributions described below; it does not establish hardware safety.

Table~\ref{tab:data_separation} records the separation between optimization, model selection, calibration, and final evaluation. The policy and its action experts are trained only on the official Human300 demonstrations. Auxiliary trajectories for the world model and risk head are generated from training-side tasks and scenes and are accounted for separately from Human300 policy training. Splits are assigned by task--scene--initialization group before rollout generation, preventing near-duplicate trajectories from crossing partitions. Neither Composite-Unseen tasks nor target-suite scenes are used to train the policy, world model, or risk head.

\begin{table*}[!tbp]
\centering
\caption{Roles of the data partitions. The official clean benchmark aggregate is never mixed with the controlled perturbation study.}
\label{tab:data_separation}
\small
\setlength{\tabcolsep}{5pt}
\begin{tabular}{@{}p{0.17\textwidth}p{0.25\textwidth}p{0.27\textwidth}p{0.22\textwidth}@{}}
\toprule
Partition & Allowed data & Permitted use & Leakage guard \\
\midrule
Policy training & Official Human300 demonstrations & Optimize the VLA and decoupled action experts & No target-suite trajectory or injected perturbation \\
Monitor training & Training-side nominal and physics-perturbed rollouts & Optimize the rolling predictor and causal risk head & No validation, calibration, or target group \\
Validation & Held-out training-side task--scene groups & Select discrepancy, span, window, suppression, guidance, and periodic-replan settings & No gradient update and no final conformal quantile \\
Conformal calibration & Nominal successful shadow-mode episodes & Fit span statistics, temporal normalization, and the functional conformal quantile & Disjoint from validation and both locked tests \\
Locked clean test & Official 50-task target suite & Report A-S, C-S, C-U, and task-weighted success & Evaluation only; no post-test tuning \\
Locked perturbation test & Preregistered composite-task subset with physical perturbations & Detector, fixed-trigger repair, latency, and failure analyses & Perturbation schedule fixed before unblinding variants \\
\bottomrule
\end{tabular}
\end{table*}

The perturbation study uses the 12 composite tasks visualized in Figures~\ref{fig:task_viz_mobile}--\ref{fig:task_viz_counter}. It balances object or base impulses, transient joint offsets, and moved-object interventions across navigation, pre-contact, post-grasp transport, and appliance-interaction phases. Paired variants share the task, scene, initial state, perturbation type and onset, environment seed, and policy-sampling seed. Detector-only comparisons replay the same shadow-mode trajectories. Rewrite comparisons restore the same simulator snapshot at the first trigger and reuse common random numbers, so they change the repair rule but neither the evidence available at the trigger nor the pre-trigger history.

The allocation uses 3 independently trained seeds, 10 clean episodes per task and seed, and 10 paired perturbation episodes per task, perturbation family, and seed, with severity and onset phase balanced within each block. Three nominal pools serve distinct roles and never mix: 400 shadow-mode nominal successes per training seed, generated by that seed's own policy and monitor, fit the conformal calibration---each checkpoint is calibrated independently, so exchangeability is with respect to deployment of the same checkpoint---a disjoint pool of 50 nominal shadow-mode episodes per perturbation-study task and seed (1{,}800 episodes) estimates empirical episode FWER, and the 1{,}500-episode locked clean suite reports benchmark success only. Table~\ref{tab:per_seed} breaks the full system's headline metrics down by training seed. Task-success uncertainty is computed by a 95\% hierarchical paired bootstrap over tasks and then episodes. Episode FWER uses an exact binomial interval, and paired success changes use an exact McNemar test. Hyperparameters are selected once on validation data; every detector variant is then independently recalibrated at target episode-level $\alpha=0.05$ before the locked test is read.

\begin{table}[!htbp]
\centering
\caption{Per-seed results for full CheckVLA. Mean matches the aggregates reported in Tables~\ref{tab:core_ablation} and~\ref{tab:trigger_comparison}.}
\label{tab:per_seed}
\footnotesize
\setlength{\tabcolsep}{2.5pt}
\begin{tabular*}{\columnwidth}{@{\extracolsep{\fill}}lcccc@{}}
\toprule
Seed & \shortstack{Clean\\Average (\%)} & \shortstack{Perturbed\\success (\%)} & \shortstack{Timely\\recall (\%)} & \shortstack{Episode\\FWER (\%)} \\
\midrule
Seed 1 & 35.6 & 48.2 & 77.1 & 5.2 \\
Seed 2 & 36.9 & 46.9 & 79.0 & 4.5 \\
Seed 3 & 35.8 & 47.7 & 77.6 & 4.7 \\
\midrule
Mean & 36.1 & 47.6 & 77.9 & 4.8 \\
\bottomrule
\end{tabular*}
\end{table}

\section{Metric Definitions}
\label{sec:supp_metrics}

Let $\mathcal{N}$ denote shadow-mode episodes that succeed under nominal execution, $T_i$ their horizons, and $t_i^*=\inf\{t:r_{i,t}>\delta_{i,t}\}$ the first alarm ($t_i^*=\infty$ when no alarm occurs). The empirical episode-level family-wise error rate (episode FWER) is
\begin{equation}
\widehat{\mathrm{FWER}}_{\mathrm{ep}}
=\frac{1}{|\mathcal{N}|}\sum_{i\in\mathcal{N}}
\mathbb{1}\!\left[\exists t\leq T_i:r_{i,t}>\delta_{i,t}\right]. \label{eq:supp_fwer}
\end{equation}
This is the probability of at least one unnecessary \emph{first} intervention on an episode that would otherwise succeed; a step-wise false-positive rate is not a substitute because it ignores repeated testing over variable horizons.

For each disrupted reference episode $i$ that fails without intervention, $\tau_{\mathrm{on},i}$ is the annotated onset of the observable deviation and $\tau_{\mathrm{irrev},i}$ is the earliest time after which no tested suffix repair can restore task success. Only the onset is a human annotation; $\tau_{\mathrm{irrev},i}$ is operational, fixed by repair outcomes under the tested rewrite rules, so timeliness is not judged against an annotator's opinion of recoverability. (In the natural-execution audit, where deviations are not injected, blinded annotators mark both times as described there.) We call the interval before $\tau_{\mathrm{irrev},i}$ the episode's \emph{empirical recovery window}; irreversibility always refers to the close of this window under the tested repair set, not to physical impossibility. A trigger is timely only if it occurs no earlier than onset and its repaired chunk can deploy before irreversibility. Timely recall and effective lead are therefore
\begin{align}
\widehat{\mathrm{Recall}}_{\mathrm{timely}}
&=\frac{1}{|\mathcal{F}|}\sum_{i\in\mathcal{F}}
\mathbb{1}\!\left[\substack{\tau_{\mathrm{on},i}\leq t_i^*,\\ t_i^*+d_{\mathrm{lat},i}<\tau_{\mathrm{irrev},i}}\right], \label{eq:supp_timely_recall}\\
\ell_i&=\tau_{\mathrm{irrev},i}-\left(t_i^*+d_{\mathrm{lat},i}\right), \label{eq:supp_effective_lead}
\end{align}
where $\mathcal{F}$ contains the disrupted reference episodes. We report the median and interquartile range of $\ell_i$ among timely alarms and retain premature, missed, and late alarms as failures in Eq.~\ref{eq:supp_timely_recall}. Premature-alarm frequency, $|\mathcal{F}|^{-1}\sum_i\mathbb{1}[t_i^*<\tau_{\mathrm{on},i}]$, is reported separately.

Repair quality is measured with paired counterfactual outcomes. Let $y_i^0$ be success from the stored trigger snapshot under the reference continuation and $y_i^{m}$ success under repair method $m$. Conditional rescue and harm rates are
\begin{align}
\mathrm{Rescue}(m)&=
\frac{\sum_i\mathbb{1}[y_i^0=0\wedge y_i^{m}=1]}
{\sum_i\mathbb{1}[y_i^0=0]}, \label{eq:supp_rescue}\\
\mathrm{Harm}(m)&=
\frac{\sum_i\mathbb{1}[y_i^0=1\wedge y_i^{m}=0]}
{\sum_i\mathbb{1}[y_i^0=1]}. \label{eq:supp_harm}
\end{align}
Rescue and harm isolate a single repair: the branch executes rule $m$ once with subsequent triggers disabled. Perturbed success instead continues the full closed loop from the same snapshot with re-intervention enabled (audited in Table~\ref{tab:reintervention}), so a rule can gain more in perturbed success than its one-shot rescue--harm gap alone implies. We additionally report task success, policy invocations per episode, and the switch discontinuity $\lVert a_{\mathrm{new},d_{\mathrm{lat}}}-a_{\mathrm{old},j+d_{\mathrm{lat}}}\rVert_2$. Together, rescue and harm distinguish useful recovery from indiscriminate replanning that merely trades one set of failures for another.

\section{Action-Conditioned Detector Ablations}
\label{sec:supp_detector}

We isolate the verifier with the action backbone, downstream rewriter, test trajectories, and target episode FWER held fixed. Each detector variant is retrained when applicable instead of being disabled at test time, and its discrepancy statistics and threshold are refit on its own calibration split. The action-shuffled row is a training-time negative control: it preserves action-token marginals and architecture while breaking the correspondence between a committed action and its predicted consequence. The constant-threshold row replaces the time-varying threshold of the main method with a single validation-tuned scalar; unlike every other row, it is not conformally calibrated, so it measures the combined effect of removing the functional shape and the conformal calibration.

The cross-family comparison additionally evaluates four policy-side signals. For the Monte Carlo entropy proxy and sampling disagreement, the policy draws the same $Q$ normalized action chunks per decision; entropy is estimated with a diagonal-Gaussian approximation, whereas disagreement is their mean pairwise squared distance. Denoising/flow-path variance is computed from clean-action estimates at validation-selected integration times. The frozen-policy feature probe is a two-layer causal classifier trained on policy features without updating the VLA. The primary comparison evaluates all policy signals at commit time. Observation-refreshed variants rerun the required policy computation after each new observation and report the resulting compute separately (Table~\ref{tab:policy_refresh}). An average episode lasts 390 control steps at 10\,Hz, so observation-rate refreshing adds 382.2 policy evaluations per episode, and its per-score p95 latency exceeds the 100\,ms control period; these variants are offline upper bounds, not deployable monitors.

\begin{table*}[!tbp]
\centering
\caption{Detector comparison at a common target episode FWER of $\alpha=0.05$. Each detector is calibrated independently on nominal successful episodes and evaluated on the same shadow trajectories. Event AUPRC treats each injected deviation as one event; effective lead includes deployment latency, and perturbed success uses the same downstream rewriter for every detector.}
\label{tab:detector_ablation}
\small
\setlength{\tabcolsep}{4.5pt}
\begin{tabular}{@{}p{0.31\textwidth}ccccc@{}}
\toprule
Detector variant & \shortstack{Episode FWER \\(\%)} & \shortstack{Event \\AUPRC $\uparrow$} & \shortstack{Timely recall \\(\%) $\uparrow$} & \shortstack{Effective lead \\(steps) $\uparrow$} & \shortstack{Perturbed success \\(\%) $\uparrow$} \\
\midrule
 
\multicolumn{6}{@{}l}{\textit{Panel A: Cross-family detector comparison}}\\
\midrule
Policy MC action-entropy proxy, commit-time
    &  5.0 &  0.512 &  58.7 &  7.8 &  38.9 \\
Policy denoising/flow-path variance, commit-time
    &  5.2 &  0.548 &  61.4 &  8.4 &  40.2 \\
Policy sampling disagreement, commit-time
    &  4.9 &  0.586 &  64.8 &  9.1 &  41.5 \\
Frozen-policy feature probe, commit-time
    &  5.1 &  0.621 &  68.6 &  10.0 &  43.7 \\
Observation-only world predictor &5.3 &0.442 &48.6 &6.1 &35.2 \\
Action-shuffled world model, retrained &5.1 &0.318 &37.9 &4.2 &31.0 \\
\textbf{Full action-conditioned verifier} & \textbf{4.8} & \textbf{0.764} & \textbf{77.9} & \textbf{13.2} & \textbf{47.6} \\
\midrule
\multicolumn{6}{@{}l}{\textit{Panel B: Internal verifier mechanisms}}\\
\midrule
   
Observation-only world predictor &5.3 &0.442 &48.6 &6.1 &35.2 \\
Action-shuffled world model, retrained &5.1 &0.318 &37.9 &4.2 &31.0 \\
  
Action conditioned, no rolling re-anchor &5.4 &0.561 &62.8 &9.2 &38.7 \\
Rolling predictor, instantaneous score &5.0 &0.634 &66.1 &10.1 &41.2 \\
Causal risk head, constant threshold &7.1 &0.701 &71.5 &11.4 &45.0 \\
\textbf{Full: time-varying conformal threshold} & \textbf{4.8} & \textbf{0.764} & \textbf{77.9} & \textbf{13.2} & \textbf{47.6} \\
\bottomrule
\end{tabular}
\end{table*}

The verifier-control rows in Table~\ref{tab:detector_ablation} separate alarm calibration from alarm usefulness. Shuffling action--future training pairs sharply degrades event ranking and timeliness. Rolling re-anchoring recovers lead time relative to a full-chunk rollout, and temporal aggregation improves over an instantaneous residual. The time-varying conformal rule improves timely recall while bringing empirical episode FWER back to the prescribed operating point, unlike the validation-tuned constant threshold. Among the policy-side rows, the frozen feature probe is strongest, yet the full verifier improves timely recall by 9.3 points and perturbed success by 3.9 points at a comparable target FWER.

\begin{table*}[!tbp]
\centering
\caption{Sensitivity to refreshing policy-side signals after new observations. Commit-time variants reuse the score produced with the dispatched chunk. Observation-refreshed variants rerun the required policy computation at the monitoring rate, with the added VLA evaluations and p95 score latency reported explicitly.}
\label{tab:policy_refresh}
\footnotesize
\setlength{\tabcolsep}{2.4pt}
\begin{tabular}{@{}p{0.22\textwidth}p{0.17\textwidth}ccccc@{}}
\toprule
Signal & Update mode & \shortstack{Extra VLA evals\\/ ep.} & \shortstack{Score p95\\(ms)} & \shortstack{Episode FWER\\(\%)} & \shortstack{Timely recall\\(\%) $\uparrow$} & \shortstack{Perturbed success\\(\%) $\uparrow$} \\
\midrule
MC action-entropy proxy & Commit-time &  0.0 &  338 &  5.0 &  58.7 &  38.9 \\
MC action-entropy proxy & Observation-refreshed &  382.2 &  338 &  5.2 &  69.5 &  44.0 \\
Sampling disagreement & Commit-time &  0.0 &  338 &  4.9 &  64.8 &  41.5 \\
Sampling disagreement & Observation-refreshed &  382.2 &  338 &  5.1 &  72.3 &  45.0 \\
Frozen policy-feature probe & Commit-time &  0.0 &  6.8 &  5.1 &  68.6 &  43.7 \\
Frozen policy-feature probe & Observation-refreshed &  382.2 &  344 &  5.0 &  73.1 &  45.4 \\
\textbf{Full action-conditioned verifier} & Observation-rate monitor &  0.0 &  16.4 & \textbf{4.8} & \textbf{77.9} & \textbf{47.6} \\
\bottomrule
\end{tabular}
\end{table*}

\section{Inference-Time Action--Consequence Binding}
\label{sec:supp_action_binding}

The training-time shuffled control does not by itself show that the trained predictor uses its action input at inference. We therefore freeze the full world model and restore held-out simulator snapshots. From each snapshot, we execute the factual action and four physically plausible counterfactuals under a common physics seed: a circular temporal shift, a gripper-state flip, a phase-matched mobility replacement, and a phase-matched manipulation replacement. Each replacement receives a consistent nominal proprioceptive rollout. For prediction $u$ and realized future $v$, we compute the span-averaged feature distance $D(u,v)$. The assignment margin compares diagonal matching, $D(0,0)+D(v,v)$, with the crossed assignment, $D(0,v)+D(v,0)$. Positive margins and pairwise accuracy above 50\% in Table~\ref{tab:action_binding} indicate action--future binding, not observation-only sensitivity. Both the action tokens and the nominal proprioceptive rollout $\hat s$ (itself a deterministic function of the actions) carry action information, so the audit certifies binding of the combined action conditioning without separating these two pathways.

\begin{table*}[!tbp]
\centering
\caption{Inference-time action--consequence binding of the frozen world model. Each candidate is generated from the same simulator snapshot and rolled out under a common physics seed. Accuracy is the fraction of pairs whose diagonal action--future assignment has lower cost than the crossed assignment.}
\label{tab:action_binding}
\small
\setlength{\tabcolsep}{4.5pt}
\begin{tabular}{@{}p{0.29\textwidth}ccccc@{}}
\toprule
Counterfactual action & \shortstack{Anchor\\pairs} & \shortstack{Predicted-future\\separation $\uparrow$} & \shortstack{Realized-future\\separation $\uparrow$} & \shortstack{Assignment\\margin $\uparrow$} & \shortstack{2AFC binding\\accuracy (\%) $\uparrow$} \\
\midrule
Within-chunk circular temporal shift &  720 &  0.214 &  0.237 &  0.118 &  78.6 \\
Gripper open/close flip &  412 &  0.181 &  0.205 &  0.094 &  74.8 \\
Phase-matched mobility-action replacement &  720 &  0.268 &  0.301 &  0.143 &  82.1 \\
Phase-matched manipulation-action replacement &  720 &  0.246 &  0.279 &  0.131 &  80.4 \\
\midrule
\textbf{Pooled counterfactuals} &  2572 &  0.233 &  0.262 &  0.125 &  79.5 \\
\bottomrule
\end{tabular}
\end{table*}

\section{Fixed-Trigger Rewriting and Latency}
\label{sec:supp_rewrite}

To prevent detector quality from confounding repair quality, every row in Table~\ref{tab:rewrite_latency} starts from the same first-trigger simulator snapshots. Every one of the 1{,}080 injected episodes produces at least one threshold crossing, so the branch set covers the full cohort; the online deployed-rewrite rate in Table~\ref{tab:reintervention} (86.2\%) is lower because online deployment additionally respects the trigger-suppression window and the empty-suffix rule. The flagged chunk, standardized exceedance, observation, episodic bank, and policy-sampling noise are shared across variants. The boundary baseline ignores the trigger until the next regular call. The unconstrained baseline generates a complete chunk and discards its already elapsed prefix only after sampling; in contrast, hard-prefix variants overwrite the irreversible prefix at every flow-integration step. Because the overwritten values do not match the training-time noise level at intermediate integration times, the clamp is an empirical projection rather than exact sampling from the conditional suffix distribution; the RTC~\cite{rtc} row provides the principled prefix-preserving inpainting alternative under the same trigger, and the switch-jump column reports the residual discontinuity of each rule. The shuffled-exceedance control preserves the marginal distribution of repair strengths but breaks its pairing with the current deviation. The validation-tuned fixed weight is $W{=}0.5$.

\begin{table*}[!tbp]
\centering
\caption{Fixed-trigger repair comparison. Rescue, harm, and switch jump are one-shot paired outcomes at the nominal latency of three control steps, with subsequent triggers disabled; perturbed success continues the full closed loop from the same snapshot with re-intervention enabled. The latency sweep varies the executable suffix while all methods start from the same first-trigger snapshots.}
\label{tab:rewrite_latency}
 
\footnotesize
\setlength{\tabcolsep}{2.8pt}
\begin{tabular}{@{}p{0.265\textwidth}cccccccc@{}}
\toprule
& & & & \multicolumn{5}{c}{Perturbed success (\%) $\uparrow$} \\
\cmidrule(lr){5-9}
Rewrite rule & \shortstack{Rescue \\(\%) $\uparrow$} & \shortstack{Harm \\(\%) $\downarrow$} & \shortstack{Switch \\jump $\downarrow$} & $d_{\mathrm{lat}}{=}0$ & $d_{\mathrm{lat}}{=}1$ & $d_{\mathrm{lat}}{=}3$ & $d_{\mathrm{lat}}{=}5$ & $d_{\mathrm{lat}}{=}10$ \\
\midrule
Wait for next chunk boundary &7.4 &4.8 & -- & 34.0 &34.0 &34.0 &34.0 & 34.0 \\
Unconstrained generation, discard prefix &10.1 &8.9 &0.284 & 48.9 &46.0 &39.8 &33.2 & 23.4 \\
Hard prefix, full rewrite ($W{=}0$) &13.2 &6.8 &0.171 & 49.5 &47.0 &42.4 &36.0 & 25.8 \\
RTC prefix-preserving inpainting & 13.7 & 4.1 & 0.091 & 53.0 & 50.6 & 45.6 & 39.7 & 29.1 \\
Hard prefix, validation-tuned fixed $W$ &12.8 &3.7 &0.104 & 53.5 &51.0 &46.0 &39.9 & 29.6 \\
Hard prefix, adaptive $W$, shuffled exceedance &14.4 &3.6 &0.103 & 51.9 &49.2 &44.3 &38.5 & 27.6 \\
\textbf{Hard prefix, adaptive $W(e)$} & \textbf{16.9} & \textbf{2.8} & \textbf{0.072} & \textbf{ 55.8} & \textbf{53.2} & \textbf{47.6} & \textbf{42.0} & \textbf{ 31.4} \\
\bottomrule
\end{tabular}
   
\end{table*}

\begin{figure}[!htbp]
\centering
\includegraphics[width=\columnwidth]{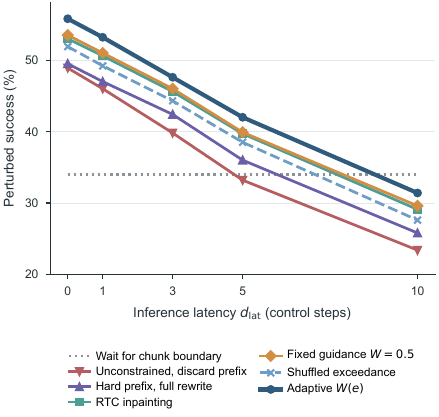}
\caption{Perturbed success across inference latency for the fixed-trigger repair rules of Table~\ref{tab:rewrite_latency}. All rules branch from the same first-trigger snapshots; the boundary baseline ignores the trigger and is latency-independent.}
\label{fig:supp_latency}
\end{figure}

The paired results attribute different roles to the two repair constraints. Hard prefixing reduces the deployment discontinuity and harm relative to post-hoc prefix disposal, whereas a full rewrite with $W{=}0$ discards all reference guidance and over-corrects otherwise recoverable trajectories. A fixed guidance weight improves this trade-off, but shuffling exceedance produces little additional gain; only the correctly paired $W(e)$ improves rescue while lowering harm.   Under the shared trigger, adaptive guidance improves nominal-latency success by  2.0 points over RTC and lowers harm by  1.3 points.   Performance decreases with inference latency for every in-chunk method because a longer irreversible prefix leaves fewer executable actions to change. The full method retains the largest advantage at the longest tested latency, but the remaining decline makes clear that verification cannot recover an episode after the useful intervention window has closed. Figure~\ref{fig:supp_latency} plots the sweep: every in-chunk method declines as the irreversible prefix grows, and adaptive guidance keeps the largest margin at every tested latency.

\section{Exceedance-to-Retention Mechanism}
\label{sec:supp_exceedance_bins}

\begin{figure}[!htbp]
\centering
\includegraphics[width=\columnwidth]{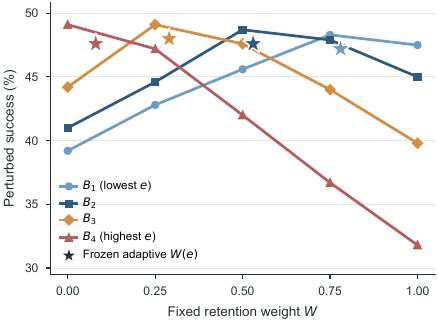}
\caption{Perturbed success under fixed retention weights, evaluated separately in each exceedance bin of Table~\ref{tab:exceedance_bins}. Stars mark the frozen adaptive mapping $W(e)$ at each bin's median exceedance.}
\label{fig:supp_exceedance}
\end{figure}

To test the mapping from standardized exceedance to reference retention, we partition positive exceedances into four equal-frequency bins using validation data and freeze the bin edges before evaluating the locked snapshots. Every snapshot (all 1{,}080 first-crossing snapshots, 270 per bin) is branched with the same policy noise under $W\in\{0,0.25,0.5,0.75,1\}$ and under the frozen adaptive mapping. The validation-selected fixed weight decreases across exceedance bins, while the test set is used only to evaluate the frozen choices. Averaged over bins, the best single fixed weight attains 46.0\% perturbed success, the frozen continuous mapping 47.6\%, and bin-wise validation-selected weights 48.8\%: exceedance-conditioned retention carries the gain, the continuous mapping captures most of it without bin edges or per-bin tuning, and the shift of the per-bin optimum toward smaller weights at larger exceedance is the monotonicity that $w_0(e)$ encodes. Figure~\ref{fig:supp_exceedance} plots the same grid: the best fixed weight shifts from 0.75 toward 0 as the exceedance bin rises, and the frozen adaptive mapping stays within 1.5 points of the per-bin envelope without bin-specific tuning.

\begin{table*}[!tbp]
\centering
\caption{Exceedance-conditioned reference retention. Panel A reports perturbed success from common first-trigger snapshots under a fixed weight grid and the frozen adaptive mapping. Panel B summarizes the validation-selected weight and adaptive repair outcomes.}
\label{tab:exceedance_bins}
\footnotesize
\setlength{\tabcolsep}{3.0pt}
\begin{tabular}{@{}p{0.17\textwidth}cccccccc@{}}
\toprule
& & \multicolumn{5}{c}{Perturbed success under fixed $W$ (\%) $\uparrow$} & \multicolumn{2}{c}{Frozen mappings} \\
\cmidrule(lr){3-7}\cmidrule(lr){8-9}
Exceedance bin & Triggers & $W{=}0$ & $W{=}0.25$ & $W{=}0.50$ & $W{=}0.75$ & $W{=}1$ & Adaptive $W(e)$ & $W_{\mathrm{val}}$ \\
\midrule
$B_1$ (lowest) &  270 &  39.2 &  42.8 &  45.6 &  48.3 &  47.5 &  47.2 &  0.75 \\
$B_2$ &  270 &  41.0 &  44.6 &  48.7 &  47.9 &  45.0 &  47.6 &  0.50 \\
$B_3$ &  270 &  44.2 &  49.1 &  47.6 &  44.0 &  39.8 &  48.0 &  0.25 \\
$B_4$ (highest) &  270 &  49.1 &  47.2 &  42.0 &  36.7 &  31.8 &  47.6 &  0.00 \\
\bottomrule
\end{tabular}

\vspace{4pt}
\begin{tabular}{@{}p{0.17\textwidth}cccccc@{}}
\toprule
Exceedance bin & \shortstack{Median\\$e$} & $W_{\mathrm{val}}$ & \shortstack{Success at\\$W_{\mathrm{val}}$ (\%)} & \shortstack{Median adaptive\\$W(e)$} & \shortstack{Adaptive rescue\\(\%) $\uparrow$} & \shortstack{Adaptive harm\\(\%) $\downarrow$} \\
\midrule
$B_1$ &  0.18 &  0.75 &  48.3 &  0.78 &  12.4 &  2.3 \\
$B_2$ &  0.46 &  0.50 &  48.7 &  0.53 &  15.8 &  2.5 \\
$B_3$ &  0.91 &  0.25 &  49.1 &  0.29 &  18.7 &  2.8 \\
$B_4$ &  1.74 &  0.00 &  49.1 &  0.08 &  23.1 &  3.2 \\
\bottomrule
\end{tabular}
\end{table*}

\section{Natural Failures and Distribution Shifts}
\label{sec:supp_external_validity}

The controlled perturbation study provides known onsets and paired snapshots, but it does not establish performance on naturally occurring failures or benign observation shifts. We therefore evaluate a held-out natural-execution cohort drawn from training-side task--scene groups excluded from monitor training and validation. Two annotators, blinded to detector scores, mark the first observable deviation, last recoverable time, failure type, and whether a nonempty suffix could plausibly repair the episode. We report both all failures and the actionable subset, without excluding failures that the policy cannot repair. The audit set is case--control: it contains every annotated failure and a matched sample of natural successes, so no success rate is defined on it. The execution-mode panel of Table~\ref{tab:natural_failure} instead re-executes the same 606 task--scene--initialization configurations under each mode with fresh randomness, so its success rates estimate the population rate rather than the case--control composition.

\begin{table*}[!tbp]
\centering
\caption{Held-out natural-failure audit; natural executions contain no injected perturbation, and onset and last recoverable time are annotated blind to detector output. The upper panel's audit set is case--control (all annotated failures plus matched success controls), so no success rate is defined on it; the lower panel re-executes the same configurations under each execution mode with fresh randomness.}
\label{tab:natural_failure}
\small
\setlength{\tabcolsep}{4.5pt}
\begin{tabular}{@{}p{0.31\textwidth}ccccc@{}}
\toprule
Evaluation subset & \shortstack{Failure\\episodes} & \shortstack{Success\\controls} & \shortstack{Event\\AUPRC $\uparrow$} & \shortstack{Timely recall\\(\%) $\uparrow$} & \shortstack{Effective lead\\(steps) $\uparrow$} \\
\midrule
All annotated natural failures &  186 &  420 &  0.683 &  70.4 &  10.6 \\
Failures with a nonempty actionable suffix &  139 &  420 &  0.721 &  78.4 &  12.1 \\
\bottomrule
\end{tabular}

\vspace{4pt}
\begin{tabular}{@{}p{0.31\textwidth}ccccc@{}}
\toprule
Execution mode & \shortstack{Overall task\\success (\%) $\uparrow$} & \shortstack{Natural-failure\\rescue (\%) $\uparrow$} & \shortstack{Natural-success\\harm (\%) $\downarrow$} & \shortstack{Deployed rewrites\\per episode $\downarrow$} & \shortstack{Subgoal regressions\\per episode $\downarrow$} \\
\midrule
Open-loop action backbone &  61.4 & -- & -- &  0.0 &  0.74 \\
Invocation-matched periodic replanning &  64.8 &  12.3 &  4.6 &  2.3 &  0.62 \\
\textbf{Full CheckVLA} & \textbf{ 68.9} & \textbf{ 21.6} & \textbf{ 2.7} & \textbf{ 0.44} & \textbf{ 0.31} \\
\bottomrule
\end{tabular}
\end{table*}

The confidently-wrong analysis uses the strongest policy-side detector selected on validation data. Both policy uncertainty and world-model risk are centered at their independently calibrated thresholds; no median split or trigger-conditioned selection is used. Each episode is assigned at a preregistered decision window before irreversibility, and rescue uses the same fixed repair rule in all quadrants. The resulting quadrant decomposition is reported in Table~\ref{tab:confidently_wrong}: the low-uncertainty, high-risk cell concentrates 48.4\% of all failures and remains the most recoverable.

The benign-shift test keeps the nominal conformal threshold fixed. Shift families and severities are selected on validation data, and at least two held-out families are absent from the benign negatives used for risk-head training. We report all predefined injections and separately require the paired no-intervention continuation to remain successful when estimating unnecessary intervention and harm.

\begin{table*}[!tbp]
\centering
\caption{Benign-deviation stress test with the nominal threshold kept fixed. Excess intervention is measured in percentage points relative to matched nominal episodes; harm is conditioned on paired no-intervention continuations that succeed.}
\label{tab:benign_deviation}
\small
\setlength{\tabcolsep}{3.8pt}
\begin{tabular}{@{}p{0.28\textwidth}cccccc@{}}
\toprule
Benign shift & Episodes & \shortstack{Reference\\success (\%) $\uparrow$} & \shortstack{Unnecessary first\\intervention (\%) $\downarrow$} & \shortstack{Excess vs. nominal\\(points) $\downarrow$} & \shortstack{Harm\\(\%) $\downarrow$} & \shortstack{CheckVLA task\\success (\%) $\uparrow$} \\
\midrule
Small camera jitter &  360 &  98.1 &  6.1 &  1.3 &  1.1 &  97.2 \\
Brief partial occlusion &  360 &  96.9 &  7.4 &  2.6 &  1.5 &  95.8 \\
Task-irrelevant background change &  360 &  98.6 &  5.6 &  0.8 &  0.7 &  98.1 \\
Small self-correcting object/base displacement &  360 &  95.8 &  6.7 &  1.9 &  1.3 &  94.9 \\
\midrule
\textbf{Pooled benign shifts} &  1440 &  97.4 &  6.5 &  1.7 &  1.2 &  96.5 \\
\bottomrule
\end{tabular}
\end{table*}

\begin{figure*}[!tbp]
\centering
\includegraphics[width=0.92\textwidth]{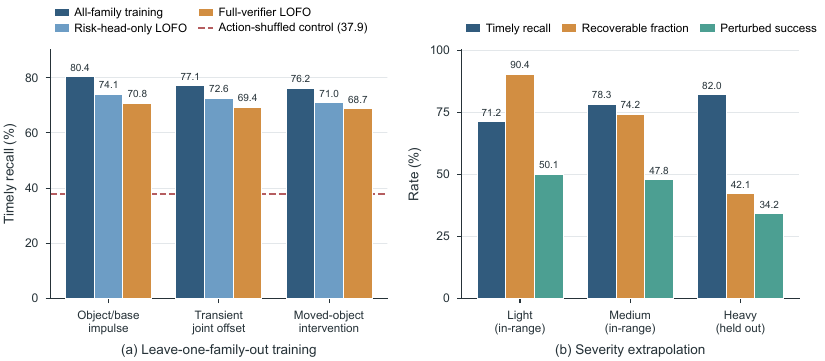}
\caption{Generalization beyond monitor-training perturbations, from Table~\ref{tab:lofo_severity}. (a) Timely recall under leave-one-family-out training; the dashed line is the action-shuffled control. (b) Severity extrapolation with the heavy range excluded from monitor training.}
\label{fig:supp_lofo}
\end{figure*}

Leave-one-family-out (LOFO) training distinguishes two scopes. Risk-head-only LOFO tests whether the head memorizes perturbation labels while retaining a world model exposed to that family. Full-verifier LOFO excludes the family from both world-model and risk-head optimization. Every changed detector is independently recalibrated on the same nominal population. Severity extrapolation is evaluated separately because high-severity events can be easier to detect but harder to repair. Figure~\ref{fig:supp_lofo} visualizes both panels: family exclusion lowers timely recall by roughly five points for the risk head and eight points for the full verifier, every held-out family stays far above the action-shuffled control, and heavier held-out severities are detected more reliably while being physically harder to repair.

\begin{table*}[!tbp]
\centering
\caption{Generalization beyond monitor-training perturbations. Panel A compares all-family training, risk-head-only LOFO, and full-verifier LOFO. Panel B evaluates severity extrapolation with the heavy range excluded from monitor training and model selection.}
\label{tab:lofo_severity}
\footnotesize
\setlength{\tabcolsep}{2.9pt}
\begin{tabular}{@{}p{0.19\textwidth}p{0.23\textwidth}ccccc@{}}
\toprule
Test family & Training exclusion & \shortstack{Episode\\FWER (\%)} & \shortstack{Event\\AUPRC $\uparrow$} & \shortstack{Timely recall\\(\%) $\uparrow$} & \shortstack{Effective lead\\(steps) $\uparrow$} & \shortstack{Perturbed success\\(\%) $\uparrow$} \\
\midrule
Object/base impulse & None; all-family reference &  4.8 &  0.781 &  80.4 &  13.8 &  49.2 \\
Object/base impulse & Risk head only &  5.1 &  0.728 &  74.1 &  12.2 &  46.1 \\
Object/base impulse & World model and risk head &  5.0 &  0.694 &  70.8 &  11.4 &  44.8 \\
Transient joint offset & None; all-family reference &  4.7 &  0.752 &  77.1 &  12.9 &  47.4 \\
Transient joint offset & Risk head only &  4.9 &  0.711 &  72.6 &  11.6 &  45.3 \\
Transient joint offset & World model and risk head &  5.2 &  0.681 &  69.4 &  10.8 &  43.9 \\
Moved-object intervention & None; all-family reference &  4.9 &  0.759 &  76.2 &  12.8 &  46.2 \\
Moved-object intervention & Risk head only &  5.0 &  0.702 &  71.0 &  11.3 &  44.0 \\
Moved-object intervention & World model and risk head &  4.8 &  0.673 &  68.7 &  10.5 &  42.6 \\
\bottomrule
\end{tabular}

\vspace{4pt}
\begin{tabular}{@{}p{0.22\textwidth}lccccc@{}}
\toprule
Test severity & Seen in monitor training? & \shortstack{Episode\\FWER (\%)} & \shortstack{Event\\AUPRC $\uparrow$} & \shortstack{Timely recall\\(\%) $\uparrow$} & \shortstack{Recoverable\\fraction (\%)} & \shortstack{Perturbed success\\(\%) $\uparrow$} \\
\midrule
Light, in-range & Yes &  4.7 &  0.721 &  71.2 &  90.4 &  50.1 \\
Medium, in-range & Yes &  4.9 &  0.772 &  78.3 &  74.2 &  47.8 \\
Heavy, held out & No &  5.1 &  0.801 &  82.0 &  42.1 &  34.2 \\
\bottomrule
\end{tabular}
\end{table*}

\section{Episodic-Memory Interaction Ablation}
\label{sec:supp_memory_ablation}

The policy reads the full keyframe bank, whereas the risk head reads only its compact summary. We therefore run a fully retrained $2\!\times\!2$ factorial ablation (Table~\ref{tab:memory_factorial}) and do not rely on test-time zeroing. When the risk-summary reader changes, its threshold is independently recalibrated; when the policy reader changes, the action model is retrained with the corresponding input path. Completed-subgoal regression counts an already satisfied task predicate that becomes unsatisfied after a replan and remains unsatisfied at episode end.

\begin{table*}[!tbp]
\centering
\caption{$2\!\times\!2$ memory ablation. A-S, C-S, C-U, and Average are clean-suite task success rates (\%); Average uses the official 18/16/16 task weighting. Every variant is retrained, and every risk-reader variant is recalibrated.}
\label{tab:memory_factorial}
\small
\setlength{\tabcolsep}{4.5pt}
\begin{tabular}{@{}llcccccc@{}}
\toprule
\shortstack{Policy \\bank} & \shortstack{Risk \\summary} & A-S & C-S & C-U & Average & \shortstack{Subgoal regressions \\/ ep. $\downarrow$} & \shortstack{Episode FWER \\(\%) $\downarrow$} \\
\midrule
No  & No  &61.5 &28.3 &9.0 &34.1 &1.34 &5.6 \\
Yes & No  &63.0 &29.7 &9.8 &35.3 &0.55 &5.5 \\
No  & Yes &62.4 &29.3 &9.4 &34.8 &1.08 &4.9 \\
\textbf{Yes} & \textbf{Yes} & \textbf{63.7} & \textbf{30.9} & \textbf{10.2} & \textbf{36.1} & \textbf{0.18} & \textbf{4.8} \\
\bottomrule
\end{tabular}
\end{table*}

The interaction distinguishes the two consumers. The policy reader accounts for most of the reduction in completed-subgoal regression and contributes most strongly on composite tasks, where evidence of earlier stations has often left the current field of view. The risk summary has a smaller direct effect on task success but brings empirical episode FWER closer to its calibrated target. Their combination gives the strongest clean-suite average and the fewest regressions, consistent with preserving both action-level progress and the context used to judge whether a discrepancy is anomalous.

To separate persistent context from storage capacity, we additionally compare memory forms under the same maximum frame budget $K$. Each policy-reader variant is retrained with its corresponding memory path, and every changed risk reader is independently recalibrated; no row is created by masking memory only at test time.

\begin{table*}[!tbp]
\centering
\caption{Memory-form comparison at a matched maximum frame budget $K$. Recent-$K$ and uniformly spaced memories control for storage capacity, whereas the event-driven bank writes frames only at validated pause-and-diversity events.}
\label{tab:memory_form}
\small
\setlength{\tabcolsep}{4pt}
\begin{tabular}{@{}p{0.25\textwidth}ccccccc@{}}
\toprule
Memory form & \shortstack{Maximum stored\\frames} & A-S & C-S & C-U & Average & \shortstack{Subgoal regressions\\per episode $\downarrow$} & \shortstack{Episode FWER\\(\%)} \\
\midrule
No persistent memory & -- &61.5 &28.3 &9.0 &34.1 &1.34 &5.6 \\
Recent-$K$ FIFO frames & $K$ & 62.3 & 29.0 & 9.5 & 34.7 & 0.92 & 5.2 \\
Uniformly spaced $K$ frames & $K$ & 62.7 & 29.4 & 9.7 & 35.1 & 0.71 & 5.0 \\
\textbf{Event-driven keyframe bank} & $K$ &\textbf{63.7} &\textbf{30.9} &\textbf{10.2} &\textbf{36.1} &\textbf{0.18} &\textbf{4.8} \\
\bottomrule
\end{tabular}
\end{table*}

\section{Calibration, Data Scale, and Repeated Interventions}
\label{sec:supp_sensitivity}

\begin{figure*}[!tbp]
\centering
\includegraphics[width=0.92\textwidth]{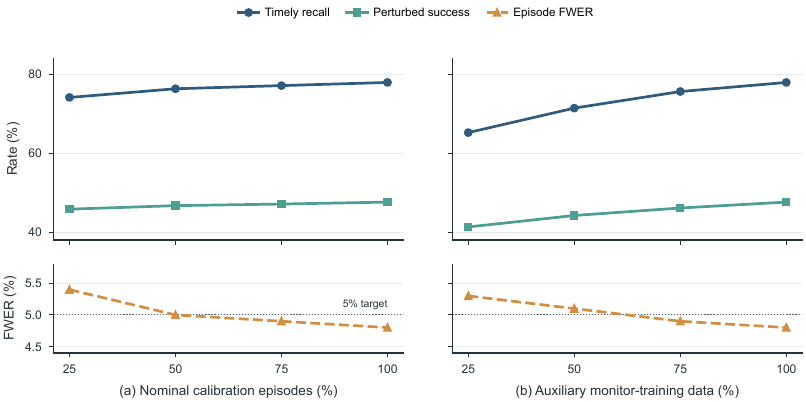}
\caption{Sensitivity to (a) the number of nominal calibration episodes and (b) the auxiliary monitor-training budget, from Tables~\ref{tab:calibration_size} and \ref{tab:monitor_data_scale}. Solid lines use the left axis; the dashed line is the empirical episode FWER on the right axis, with the dotted line at the 5\% target.}
\label{fig:supp_scaling}
\end{figure*}

Calibration-size sensitivity changes only the nominal calibration sample, not detector weights, validation choices, or locked test episodes. We repeatedly subsample the same calibration pool and retain only sizes for which the finite-sample conformal quantile is defined at $\alpha=0.05$. Monitor-data sensitivity instead retrains the world model and risk head on nested task--scene--initialization groups, followed by an independent nominal calibration for each row. Figure~\ref{fig:supp_scaling} summarizes both sensitivities: detection metrics saturate as either resource grows, while the empirical episode FWER stays near the 5\% target at every size.

\begin{table*}[!tbp]
\centering
\caption{Sensitivity to the number of nominal calibration episodes. Values for subsampled rows are averages across 20 resamples; event AUPRC is threshold-independent and is therefore unchanged because detector weights and event scores are fixed.}
\label{tab:calibration_size}
\footnotesize
\setlength{\tabcolsep}{2.9pt}
\begin{tabular}{@{}p{0.16\textwidth}cccccccc@{}}
\toprule
Calibration fraction & $n_{\mathrm{cal}}$ & Resamples & \shortstack{Finite $\hat q_\alpha$\\(\%)} & \shortstack{Episode\\FWER (\%)} & \shortstack{Event\\AUPRC} & \shortstack{Timely recall\\(\%)} & \shortstack{Effective lead\\(steps)} & \shortstack{Perturbed success\\(\%)} \\
\midrule
25\% &  100 &  20 &  100 &  5.4 &  0.764 &  74.1 &  12.1 &  45.8 \\
50\% &  200 &  20 &  100 &  5.0 &  0.764 &  76.3 &  12.7 &  46.7 \\
75\% &  300 &  20 &  100 &  4.9 &  0.764 &  77.1 &  13.0 &  47.1 \\
100\% &  400 & -- &  100 &4.8 &0.764 &77.9 &13.2 &47.6 \\
\bottomrule
\end{tabular}
\end{table*}

\begin{table*}[!tbp]
\centering
\caption{Sensitivity to auxiliary monitor-training data. Training subsets are nested by task--scene--initialization group; the action backbone is fixed, each monitor is retrained from scratch, and each row receives an independent nominal conformal calibration.}
\label{tab:monitor_data_scale}
\small
\setlength{\tabcolsep}{4.5pt}
\begin{tabular}{@{}p{0.20\textwidth}cccccc@{}}
\toprule
Monitor-training fraction & \shortstack{Training\\rollouts} & \shortstack{Episode\\FWER (\%)} & \shortstack{Event\\AUPRC $\uparrow$} & \shortstack{Timely recall\\(\%) $\uparrow$} & \shortstack{Effective lead\\(steps) $\uparrow$} & \shortstack{Perturbed success\\(\%) $\uparrow$} \\
\midrule
25\% &  900 &  5.3 &  0.638 &  65.2 &  9.8 &  41.3 \\
50\% &  1800 &  5.1 &  0.701 &  71.4 &  11.4 &  44.2 \\
75\% &  2700 &  4.9 &  0.742 &  75.6 &  12.6 &  46.1 \\
100\% &  3600 &4.8 &0.764 &77.9 &13.2 &47.6 \\
\bottomrule
\end{tabular}
\end{table*}

Benign negatives can lower false interventions under harmless shifts, but they could also suppress sensitivity to real deviations. We therefore retrain the risk head with and without benign negatives while holding the world model, policy, data budget, and nominal calibration protocol fixed (Table~\ref{tab:benign_training}). Benign negatives leave sensitivity to real deviations essentially unchanged while cutting benign first interventions from 14.8\% to 6.5\% and benign harm from 4.7\% to 1.2\%; the 1.1-point higher timely recall of the benign-free head is obtained at a 6.2\% empirical FWER, not at the target operating point. Although both variants are recalibrated on the same nominal pool, the head trained without benign negatives reacts to harmless appearance variation that the calibration pool covers only sparsely, which shifts its deployed error rate above the target; benign-aware training keeps the deployed operating point at the prescribed 4.8\%.

\begin{table*}[!tbp]
\centering
\caption{Ablation of benign negatives in risk-head training. The benign intervention and harm columns use the held-out shift families in Table~\ref{tab:benign_deviation}; detector metrics use the unchanged controlled perturbation test.}
\label{tab:benign_training}
\small
\setlength{\tabcolsep}{4.5pt}
\begin{tabular}{@{}p{0.27\textwidth}cccccc@{}}
\toprule
Risk-head training & \shortstack{Episode\\FWER (\%)} & \shortstack{Benign first\\intervention (\%) $\downarrow$} & \shortstack{Benign harm\\(\%) $\downarrow$} & \shortstack{Event\\AUPRC $\uparrow$} & \shortstack{Timely recall\\(\%) $\uparrow$} & \shortstack{Perturbed success\\(\%) $\uparrow$} \\
\midrule
Without benign negatives &  6.2 &  14.8 &  4.7 &  0.772 &  79.0 &  47.4 \\
\textbf{With benign negatives} & \textbf{4.8} & \textbf{ 6.5} & \textbf{ 1.2} &0.764 &77.9 &\textbf{47.6} \\
\bottomrule
\end{tabular}
\end{table*}

Detection performance is also decomposed by task phase in Table~\ref{tab:phase_false_alarm}, so that aggregate recall cannot hide late or weak performance around contact. Phase labels and negative nominal windows are fixed before detector outputs are read. Step-wise false-positive rate is reported only as a diagnostic alongside episode FWER, not as a replacement for repeated-testing control.

\begin{table*}[!tbp]
\centering
\caption{Task-phase and false-alarm decomposition. Panel A reports event-level detector behavior by perturbation onset phase. Panel B reports nominal false alarms at both episode and step levels under the same target episode FWER.}
\label{tab:phase_false_alarm}
\footnotesize
\setlength{\tabcolsep}{3.0pt}
\begin{tabular}{@{}p{0.23\textwidth}cccccc@{}}
\toprule
Perturbation onset phase & Events & \shortstack{Event\\AUPRC $\uparrow$} & \shortstack{Timely recall\\(\%) $\uparrow$} & \shortstack{Effective lead\\(steps) $\uparrow$} & \shortstack{Premature alarm\\(\%) $\downarrow$} & \shortstack{Perturbed success\\(\%) $\uparrow$} \\
\midrule
Navigation &  270 &  0.731 &  73.7 &  14.8 &  2.2 &  46.1 \\
Pre-contact alignment &  270 &  0.748 &  75.9 &  13.9 &  2.6 &  46.9 \\
Post-grasp transport &  270 &  0.792 &  82.2 &  12.7 &  3.0 &  49.8 \\
Appliance interaction &  270 &  0.761 &  79.6 &  11.4 &  3.3 &  47.6 \\
\bottomrule
\end{tabular}

\vspace{4pt}
\begin{tabular}{@{}p{0.27\textwidth}ccccc@{}}
\toprule
Detector & \shortstack{Nominal\\episodes} & \shortstack{Episode\\FWER (\%)} & \shortstack{False alarms per\\1000 steps} & \shortstack{Median first-alarm\\time (steps)} & \shortstack{Policy calls\\per episode} \\
\midrule
Observation-only world predictor &  1800 &5.3 & 1.31 & 84 & 10.0 \\
Action-shuffled world model &  1800 &5.1 & 1.27 & 86 & 9.8 \\
Full action-conditioned verifier &  1800 &4.8 & 1.12 & 91 &10.2 \\
\bottomrule
\end{tabular}
\end{table*}

The conformal statement covers the first intervention only. We therefore audit repeated deployment after the first repair, counting a rewrite only when its suffix is actually deployed. Re-alarms during inference or cooldown are logged separately and do not count as additional interventions.

\begin{table*}[!tbp]
\centering
\caption{Audit of repeated interventions after the first deployment. Post-repair re-alarm uses a validation-fixed window of $K_{\mathrm{audit}}$ control steps.}
\label{tab:reintervention}
\footnotesize
\setlength{\tabcolsep}{2.6pt}
\begin{tabular}{@{}p{0.22\textwidth}ccccccc@{}}
\toprule
Evaluation cohort & Episodes & \shortstack{$\geq1$ deployed\\rewrite (\%)} & \shortstack{$\geq2$ deployed\\rewrites (\%)} & \shortstack{Mean rewrites\\per episode} & \shortstack{Re-alarm within\\$K_{\mathrm{audit}}$ (\%)} & \shortstack{Post-repair subgoal\\regression (\%)} & \shortstack{Harm\\(\%) $\downarrow$} \\
\midrule
Clean target-suite episodes &  1500 &  4.8 &  0.7 &  0.06 &  8.3 &  1.1 &  2.1 \\
Benign-shift episodes &  1440 &  6.5 &  1.0 &  0.08 &  9.4 &  1.3 &  1.2 \\
Injected-perturbation episodes &  1080 &  86.2 &  16.8 &  1.04 &  14.5 &  3.2 &  2.8 \\
Held-out natural executions &  606 &  34.8 &  7.9 &  0.44 &  12.1 &  2.9 &  2.7 \\
\bottomrule
\end{tabular}
\end{table*}

\section{Runtime and Complexity}
\label{sec:supp_runtime}

Runtime is measured end to end with low-level control active, after warm-up, on the same workstation and batch size for every variant. The periodic baseline runs the verifier in shadow mode and is tuned on validation data to match the full method's mean number of VLA invocations; it therefore controls for both monitoring compute and extra policy calls. Added trainable parameters and monitor FLOPs are measured relative to the same decoupled action backbone and exclude the frozen world encoder, whose memory footprint is nevertheless included in peak VRAM.

\begin{table*}[!tbp]
\centering
\caption{Runtime and complexity comparison. The periodic baseline matches the full method's average VLA-call budget and carries the same shadow verifier.}
\label{tab:runtime}
\small
\setlength{\tabcolsep}{5pt}
\begin{tabular}{@{}p{0.23\textwidth}cccccc@{}}
\toprule
Execution mode & \shortstack{Added trainable \\params (M)} & \shortstack{Added GFLOPs \\/ step} & \shortstack{Monitor p50 / p95 \\(ms)} & \shortstack{VLA calls \\/ ep.} & \shortstack{Wall-clock \\factor} & \shortstack{Peak VRAM \\(GB)} \\
\midrule
Open-loop action backbone &0.0 &0.0 & -- &7.8 &1.00$\times$ &21.4 \\
Verifier in shadow mode &88.4 &16.9 &12.1 / 16.4 &7.8 &1.06$\times$ &23.1 \\
Invocation-matched periodic replanning &88.4 &16.9 &12.1 / 16.4 &10.1 &1.18$\times$ &23.3 \\
\textbf{Full CheckVLA} &88.4 &16.9 &12.1 / 16.4 &10.2 &1.18$\times$ &23.3 \\
\bottomrule
\end{tabular}
\end{table*}

The VLA latency is 286 / 338 ms at p50/p95 under a 10 Hz controller, corresponding to the nominal $d_{\mathrm{lat}}=3$ steps illustrated in Figure~\ref{fig:suffix_timeline}. The verifier remains below one control period at p95 and overlaps with execution. Invocation matching makes the principal comparison interpretable: a difference between periodic replanning and CheckVLA reflects how the extra calls are timed and used rather than how many are spent; the verified-trigger row of Table~\ref{tab:core_ablation} isolates the timing component alone (+3.9 points).

The fixed-trigger comparison isolates the repair sampler, whereas RTC~\cite{rtc} and REMAC~\cite{remac} also define native asynchronous execution procedures. We therefore report a separate end-to-end comparison. RTC uses its published training-free schedule; full REMAC is retrained with masked action chunks and is not represented by an inference-only sampler. Validation selects each asynchronous horizon before the locked test, and policy calls and wall-clock cost remain explicit.

\begin{table*}[!tbp]
\centering
\caption{End-to-end asynchronous execution comparison. Clean Average uses the official task weighting; perturbed success, harm, policy calls, and wall-clock cost use the controlled perturbation protocol. Training adaptation distinguishes training-free scheduling from methods that modify the action model or add a verifier.}
\label{tab:asynchronous_baselines}
\small
\setlength{\tabcolsep}{4.3pt}
\begin{tabular}{@{}p{0.25\textwidth}cccccl@{}}
\toprule
Execution mode & \shortstack{Clean\\Average (\%)} & \shortstack{Perturbed\\success (\%)} & \shortstack{VLA calls\\per episode} & \shortstack{Wall-clock\\factor} & \shortstack{Harm\\(\%)} & Training adaptation \\
\midrule
Open-loop action backbone &21.6 & 28.6 &7.8 &1.00$\times$ & -- & None \\
RTC, native asynchronous schedule & 30.2 & 40.8 & 10.1 & 1.17$\times$ & 4.0 & None \\
Full REMAC & 31.4 & 42.6 & 10.3 & 1.21$\times$ & 3.5 & Masked-chunk training \\
\textbf{Full CheckVLA} &\textbf{36.1} &\textbf{47.6} &10.2 &1.18$\times$ &\textbf{2.8} & Verifier training \\
\bottomrule
\end{tabular}
\end{table*}

\section{Failure Analysis and Evaluation Boundaries}
\label{sec:supp_failures}

A blinded annotation audit assigns one primary proximate cause to each of a sample of 140 residual failures from the full method on the perturbation test, reaching Cohen's $\kappa=0.79$ before adjudication. The taxonomy attributes 31.4\% of residual failures to a missed or late alarm that left no useful suffix, 23.6\% to a physically unrecoverable state before deployment, 18.6\% to world-model ambiguity under occlusion or contact, 14.3\% to a semantic or action-generation error that a correct trigger could not repair, 7.1\% to repair-induced state regression or repeated triggering, and 5.0\% to an unnecessary intervention on an otherwise successful continuation. These mutually exclusive shares sum to 100.0\% after adjudication.

The distribution separates two distinct limiting factors. Late alarms and long inference latency reduce the suffix that remains physically modifiable; better rewriting cannot recover this group. Conversely, policy-semantic errors and ambiguous world-model features may be detected on time but still lack a competent replacement action. This distinction motivates reporting timely recall, rescue, and harm separately; final success alone is not a detector metric. It also bounds the current claim: functional conformal calibration controls only the unnecessary \emph{first} intervention on exchangeable nominal-success episodes. It does not guarantee recall under perturbations, safety after a repair, coverage after distribution shift, or transfer from simulation to hardware. Those properties remain empirical and must be re-evaluated whenever the task distribution, observation stack, controller frequency, or risk-reader memory changes.

\section{Additional Visualizations}

\begin{figure*}[!tbp]
\centering
\includegraphics[width=0.96\textwidth]{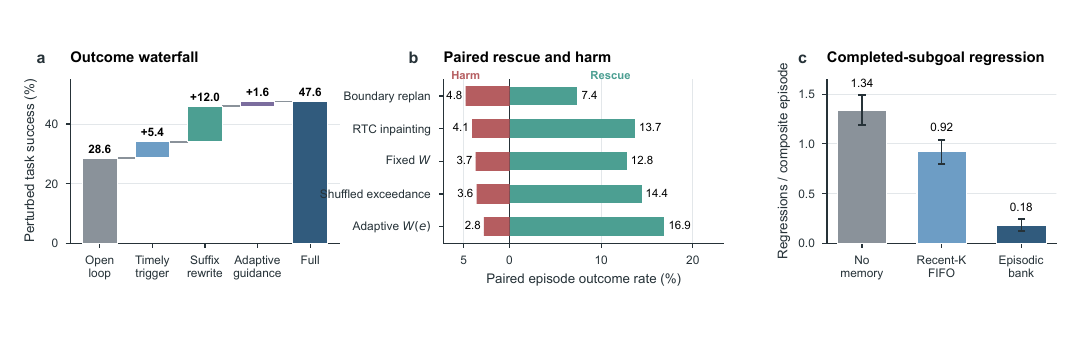}
\caption{Outcome visualization. The panels show (left) an additive waterfall of perturbed success from the open-loop backbone to full CheckVLA, (middle) paired rescue and harm rates for the fixed-trigger repair rules, and (right) completed-subgoal regression for matched memory forms.}
\label{fig:outcomes}
\end{figure*}

Figure~\ref{fig:outcomes} complements the tabular results with three outcome views. The left panel decomposes the perturbed-success difference between the open-loop backbone (28.6\%) and full CheckVLA (47.6\%) into the contributions of the timely verified trigger, the latency-aware suffix rewrite, and risk-adaptive guidance, with each level anchored to a row of Table~\ref{tab:rewrite_latency} or Table~\ref{tab:asynchronous_baselines}. The middle panel shows paired rescue and harm rates for the fixed-trigger repair rules of Table~\ref{tab:rewrite_latency}; the joint improvement of higher rescue at lower harm appears only for the correctly paired adaptive guidance. The right panel reports completed-subgoal regression for the matched memory forms of Table~\ref{tab:memory_form}, isolating the contribution of the event-driven bank to preserving prior progress.

\FloatBarrier

\section{Task Visualizations on RoboCasa365}

\begin{figure*}[p]
\centering
\includegraphics[page=1,width=\textwidth,height=0.92\textheight,keepaspectratio,trim=18 30 18 30,clip]{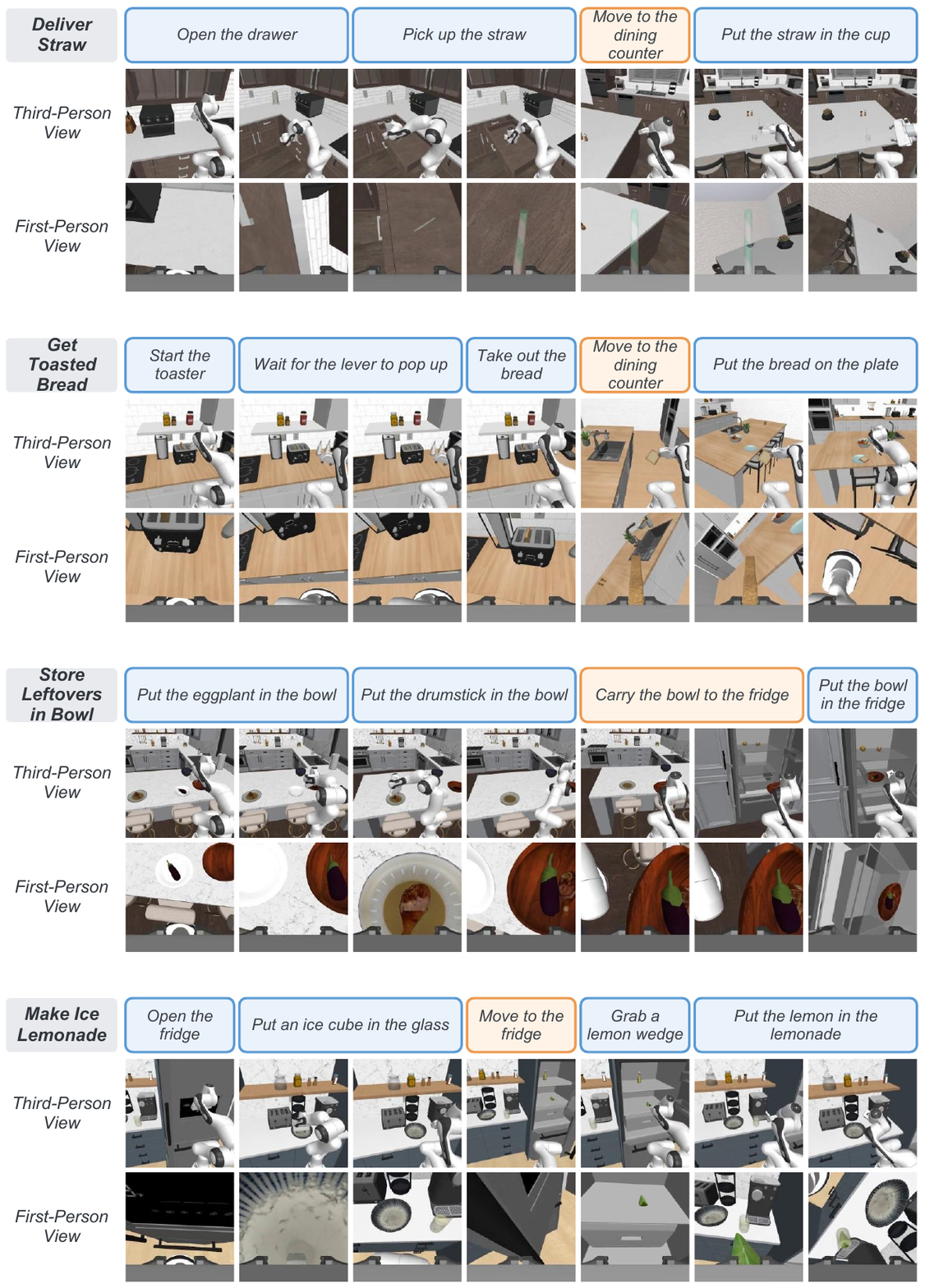}
\caption{Task visualizations on RoboCasa365 (I): cross-station delivery tasks. Blue prompt boxes mark manipulation segments and orange boxes mark base locomotion; the two rows show time-aligned third-person and egocentric wrist views. Tasks: \textit{Deliver Straw}, \textit{Get Toasted Bread}, \textit{Store Leftovers in Bowl}, \textit{Make Ice Lemonade}.}
\label{fig:task_viz_mobile}
\end{figure*}

\begin{figure*}[p]
\centering
\includegraphics[page=2,width=\textwidth,height=0.92\textheight,keepaspectratio,trim=18 30 18 30,clip]{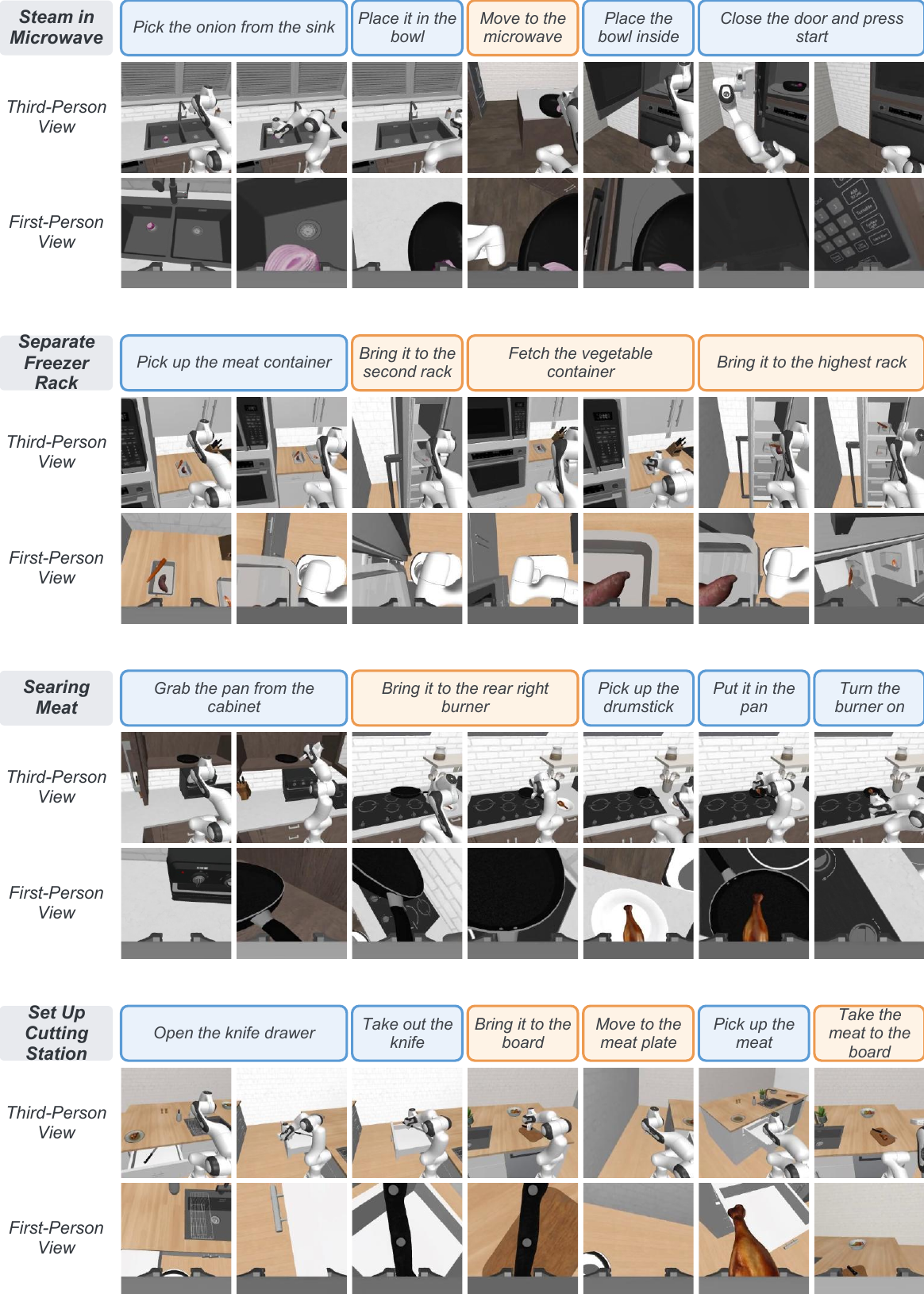}
\caption{Task visualizations on RoboCasa365 (II): tasks interleaving base locomotion with appliance operation. Format as in Figure~\ref{fig:task_viz_mobile}. Tasks: \textit{Steam in Microwave}, \textit{Separate Freezer Rack}, \textit{Searing Meat}, \textit{Set Up Cutting Station}.}
\label{fig:task_viz_appliance}
\end{figure*}

\begin{figure*}[p]
\centering
\includegraphics[page=3,width=\textwidth,height=0.92\textheight,keepaspectratio,trim=18 30 18 30,clip]{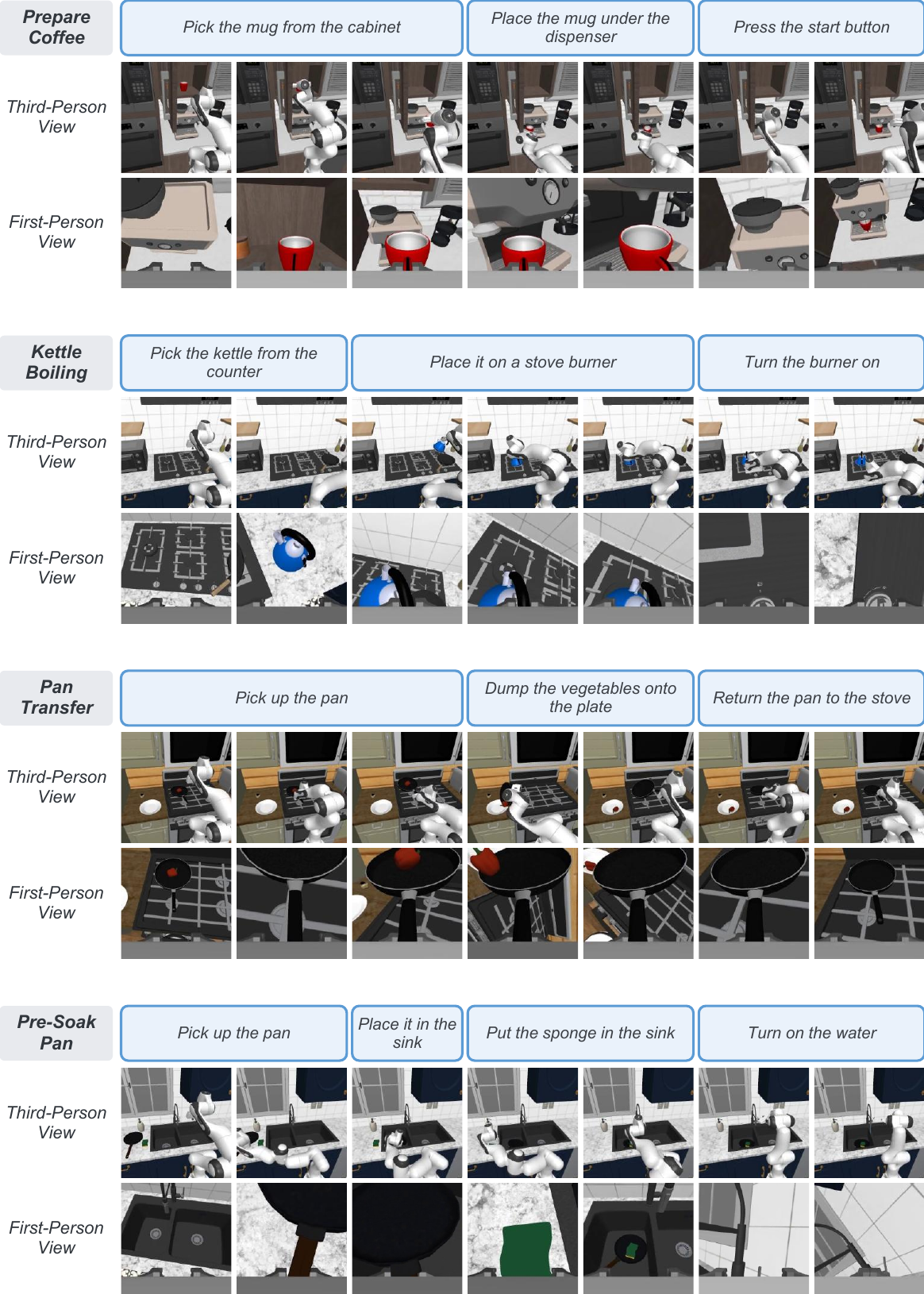}
\caption{Task visualizations on RoboCasa365 (III): stationary appliance and counter-top manipulation tasks. Format as in Figure~\ref{fig:task_viz_mobile}. Tasks: \textit{Prepare Coffee}, \textit{Kettle Boiling}, \textit{Pan Transfer}, \textit{Pre-Soak Pan}.}
\label{fig:task_viz_counter}
\end{figure*}

Figures~\ref{fig:task_viz_mobile}--\ref{fig:task_viz_counter} visualize twelve representative composite tasks from the official RoboCasa365 target benchmark. Each composite task issues a natural-language instruction that chains several sub-goals, often across different stations of the kitchen, so the mobile manipulator must interleave base locomotion with arm control: fetching an object at one station, transporting it while grasped, and operating articulated fixtures (doors, drawers, knobs, buttons) at another.

Each storyboard follows the same format. The top ribbon decomposes the episode's instruction into sub-task prompts, aligned with the keyframe columns they span; blue boxes denote manipulation segments, and orange boxes denote segments that involve base locomotion. The two rows below show seven time-aligned keyframes from the third-person scene camera and the egocentric wrist camera, the two views consumed by the policy.

% Template-native column break: the AAAI style forbids balancing packages.
\newpage
\noindent The three figures group the tasks by workspace structure: Figure~\ref{fig:task_viz_mobile} shows cross-station deliveries, whose instructions explicitly require driving between stations (e.g., taking a straw from a drawer and delivering it to a cup on the dining counter); Figure~\ref{fig:task_viz_appliance} shows tasks that interleave locomotion with appliance operation (microwave, freezer, stove, cutting station); and Figure~\ref{fig:task_viz_counter} shows stationary appliance and counter-top manipulation. Episodes of this kind chain partially irreversible sub-goals over a long horizon---switching appliances on, releasing objects into containers, closing doors on loaded racks---which is precisely the setting in which an in-chunk deviation is costly if it goes unnoticed, and which motivates the action-conditioned verification and in-chunk replanning mechanism formalized in Eqs.~\ref{eq:rollout}--\ref{eq:field}.

\FloatBarrier

% Keep the final section heading, its lead-in, and the full-width configuration
% table together; a terminal table* would otherwise strand the heading on a
% nearly empty page because double-column floats can only enter at page top.
\begin{strip}
\section{Complete Configuration}
\label{sec:supp_config}

Table~\ref{tab:config} consolidates every hyperparameter of the executed system, its training and calibration, and the baseline configurations. Validation-selected entries were fixed on the validation split of Table~\ref{tab:data_separation} before either locked test was read; no value was tuned on the target suites.

\begin{center}
\centering
\captionof{table}{Complete configuration. Validation-selected values are chosen once on the validation split before the locked tests; architecture and infrastructure entries are fixed design choices.}
\label{tab:config}
\footnotesize
\setlength{\tabcolsep}{3.5pt}
\begin{tabular}{@{}llll@{}}
\toprule
Component & Hyperparameter & Value & Selected on \\
\midrule
\multicolumn{4}{@{}l}{\textit{Execution}}\\
Chunked execution & Chunk length $H$ & 50 & Fixed a priori \\
Chunked execution & Flow-integration steps & 10 ($\Delta u{=}0.1$) & Fixed a priori \\
Chunked execution & Deployment schedule $d_{\mathrm{lat}}$ & 3 control steps & Fixed a priori \\
\midrule
\multicolumn{4}{@{}l}{\textit{Verification}}\\
Rolling prediction & Span bound $k$ & 8 & Validation \\
Discrepancy & Distance $\mathrm{dist}$ & cosine on pooled, normalized features & Validation \\
Discrepancy & Standardization floor $\varepsilon$ & $10^{-3}$ & Fixed a priori \\
Risk head & Window $w$ & 32 & Validation \\
Trigger & Target level $\alpha$ & 0.05 & Fixed a priori \\
Trigger & Suppression length $s_{\min}$ & 8 ($=k$) & Validation \\
\midrule
\multicolumn{4}{@{}l}{\textit{Suffix rewriting}}\\
Retention & Entry-scale floor $w_{\min}$ & 0.0 & Validation \\
Retention & Exceedance sensitivity $\beta$ & 1.38 & Validation \\
Positional decay & $\lambda_{\mathrm{mobility}}/\lambda_{\mathrm{manip}}$, replan & 0.60 / 0.90 & Validation \\
Positional decay & $\lambda_{\mathrm{mobility}}/\lambda_{\mathrm{manip}}$, regular & 0.10 / 0.15 & Validation \\
\midrule
\multicolumn{4}{@{}l}{\textit{Episodic memory}}\\
Keyframe bank & Budget $K$ / tokens per keyframe & 8 / 16 & Validation \\
Write filter & Pause window $w_s$ / confirm delay & 5 / 3 & Validation \\
Write filter & Min.\ write gap / similarity threshold & 10 / 0.95 & Validation \\
Fusion & Gate bias initialization & $-4$ & Fixed a priori \\
\midrule
\multicolumn{4}{@{}l}{\textit{Monitor architecture}}\\
Latent predictor $\Psi$ & Width / depth / heads & 960 / 7 / 12 & Fixed a priori \\
Risk head $R$ & Width / depth / heads & 384 / 4 / 6 & Fixed a priori \\
Monitor total & Added trainable parameters & 88.4M & -- \\
\midrule
\multicolumn{4}{@{}l}{\textit{Training and calibration}}\\
$\Psi$ optimization & AdamW: lr / batch / steps & $3{\times}10^{-4}$ / 32 / 60k TF $+$ 20k self-rollout & -- \\
$R$ optimization & AdamW: lr / batch / steps & $1{\times}10^{-4}$ / 64 / 20k & -- \\
$R$ supervision & Onset window $\Delta$ & 20 control steps & Fixed a priori \\
$R$ supervision & Mix nominal : natural : perturbed & $2{:}1{:}1$ (1{,}800 / 900 / 900 rollouts) & Fixed a priori \\
Conformal & Fit / calibration episodes per seed & 200 / 400 & Fixed a priori \\
Audits & Re-alarm window $K_{\mathrm{audit}}$ & 20 control steps & Validation \\
\midrule
\multicolumn{4}{@{}l}{\textit{Baselines}}\\
Periodic replanning & Replan interval & 39 control steps & Validation \\
Policy-side detectors & Sample count $Q$ & 8 & Fixed a priori \\
RTC / REMAC & Asynchronous horizon & 25 control steps & Validation \\
\midrule
\multicolumn{4}{@{}l}{\textit{Infrastructure}}\\
Hardware & GPU / precision & A100 80\,GB / bf16 & -- \\
Software & Framework & PyTorch 2.3.1, CUDA 12.1 & -- \\
Software & Simulator & RoboCasa365 (robosuite 1.4.1, MuJoCo 3.1.4) & -- \\
Runs & Random seeds & 1 / 2 / 3 & -- \\
\bottomrule
\end{tabular}
\end{center}
\end{strip}
% Flush the terminal strip: unlike a mid-document strip, it has no following
% prose to trigger the two-column output routine before \end{document}.
\noindent\mbox{}\par
\clearpage

\end{document}